\pdfoutput=1

\documentclass[11pt]{article}

\usepackage[preprint]{acl}

\usepackage{times}
\usepackage{latexsym}
\usepackage[T1]{fontenc}

\usepackage[utf8]{inputenc}

\usepackage{microtype}

\usepackage{inconsolata}

\usepackage{graphicx}
\usepackage{amsmath}
\usepackage{float}  
\usepackage{booktabs}
\usepackage{multirow}
\usepackage{makecell}
\usepackage{float}
\usepackage{subcaption}
\usepackage{tcolorbox}
\usepackage{booktabs} 
\usepackage{colortbl} 
\usepackage[table]{xcolor} 
\usepackage[ruled,vlined]{algorithm2e} 
\usepackage{setspace}
\DontPrintSemicolon 
\usepackage{xurl}

\title{ConCISE: Confidence-guided Compression in Step-by-step Efficient Reasoning}

\author{
Ziqing Qiao$^{1}$\thanks{Work was done when Ziqing Qiao was interning at Pattern Recognition Center, WeChat AI, Tencent Inc.}, 
Yongheng Deng$^{1}$\thanks{Corresponding authors},
Jiali Zeng$^{2}$, 
Dong Wang$^{1}$, 
Lai Wei$^{1}$, 
Guanbo Wang$^{1}$, \\
{\bf Fandong Meng}$^{2}$\footnotemark[2],   
{\bf Jie Zhou}$^{2}$, 
{\bf Ju Ren}$^{1}$, 
{\bf Yaoxue Zhang}$^{1}$ \\
$^{1}$Department of Computer Science and Technology, Tsinghua University \\
$^{2}$Pattern Recognition Center, WeChat AI, Tencent Inc., China\\
  \texttt{qzq24@mails.tsinghua.edu.cn} \\ \texttt{dengyh1013@gmail.com}\hspace{1em}
  \texttt{fandongmeng@tencent.com}
}

\begin{document}
\maketitle

\begin{abstract}
Large Reasoning Models (LRMs) perform strongly in complex reasoning tasks via Chain-of-Thought (CoT) prompting, but often suffer from verbose outputs, increasing computational overhead. Existing fine-tuning-based compression methods either operate post-hoc pruning, risking disruption to reasoning coherence, or rely on sampling-based selection, which fails to remove redundant content thoroughly. To address these limitations, this work begins by framing two key patterns of redundant reflection in LRMs—\textit{Confidence Deficit}, wherein the model reflects on correct intermediate steps, and \textit{Termination Delay}, where reflection continues after a verified, confident answer—through a confidence-guided perspective. Based on this, we introduce \textsc{ConCISE} (\textbf{Con}fidence-guided \textbf{C}ompression \textbf{I}n \textbf{S}tep-by-step \textbf{E}fficient Reasoning), a framework designed to generate concise reasoning chains, integrating \textit{Confidence Injection} to boost reasoning confidence, and \textit{Early Stopping} to terminate reasoning when confidence is sufficient. Extensive experiments demonstrate that compared to baseline methods, fine-tuning LRMs on \textsc{ConCISE}-generated data yields a better balance between compression and task performance, reducing length by up to \textasciitilde50\% under SimPO, while maintaining high task accuracy.
\end{abstract}

\section{Introduction}

In recent years, Large Language Models (LLMs) have achieved remarkable progress in natural language processing, particularly in complex reasoning tasks. Large Reasoning Models (LRMs), such as OpenAI-o1 \cite{jaech2024openai} and DeepSeek-R1 \cite{guo2025deepseek}, which leverage the Chain-of-Thought paradigm \cite{wei2022chain}, have demonstrated state-of-the-art performance on benchmarks involving mathematical problem solving and logical reasoning\cite{muennighoff2025s1, qwq-32b-preview}. However, a prominent limitation of LRMs is their tendency to generate excessively verbose reasoning chains \cite{feng2025efficient, chen2024not}, incuring considerable computational overhead, and posing challenges for deployment in resource-constrained settings \cite{team2025kimi}.
\begin{figure}[t]
\centering
\includegraphics[width=\linewidth]{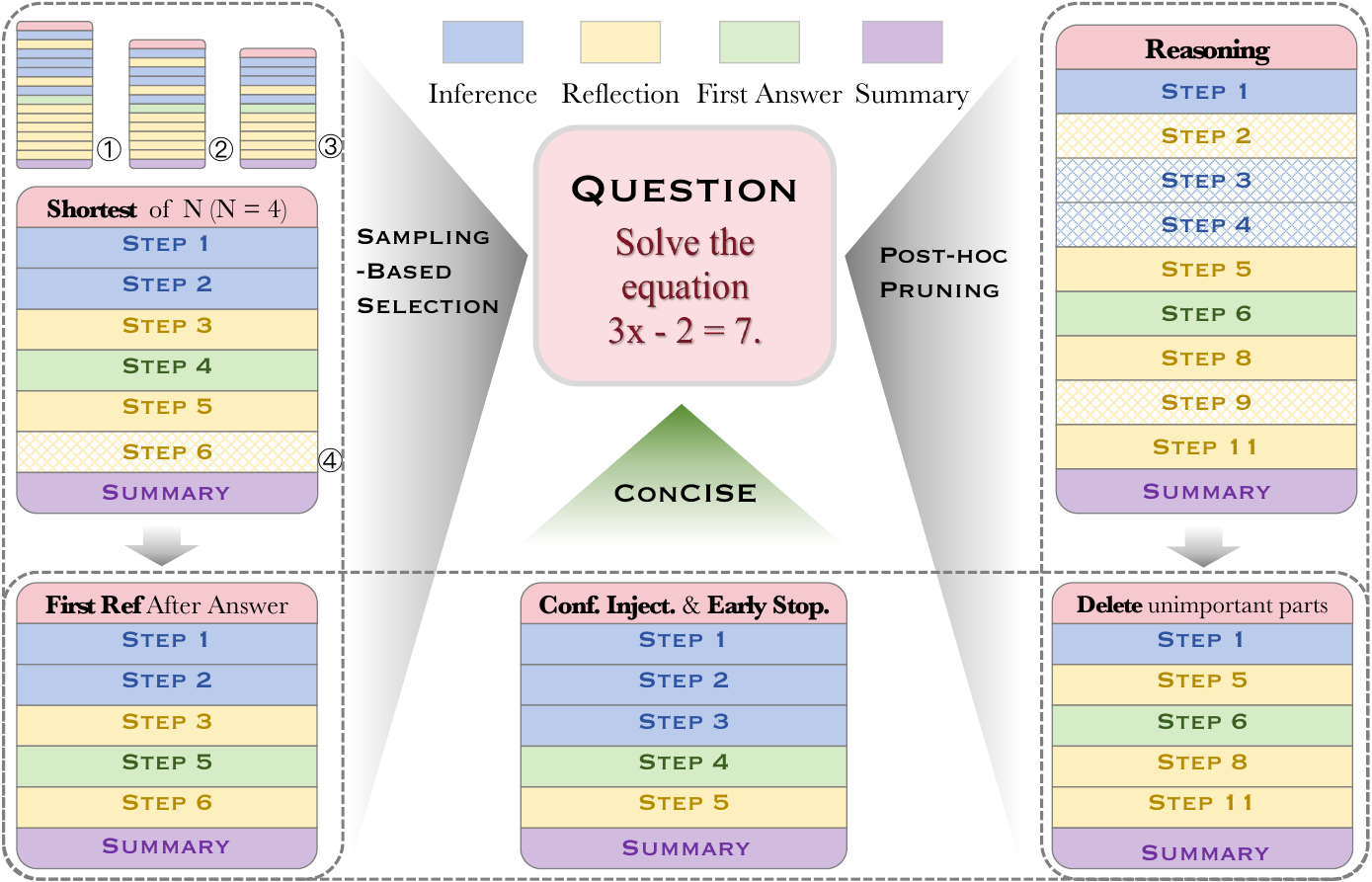}
\caption{Training dataset construction workflows: \textsc{ConCISE} (our proposed method) vs. existing methods.}
\label{fig:Comparison}
\end{figure}
\begin{figure}[t]
\centering
\includegraphics[width=\linewidth]{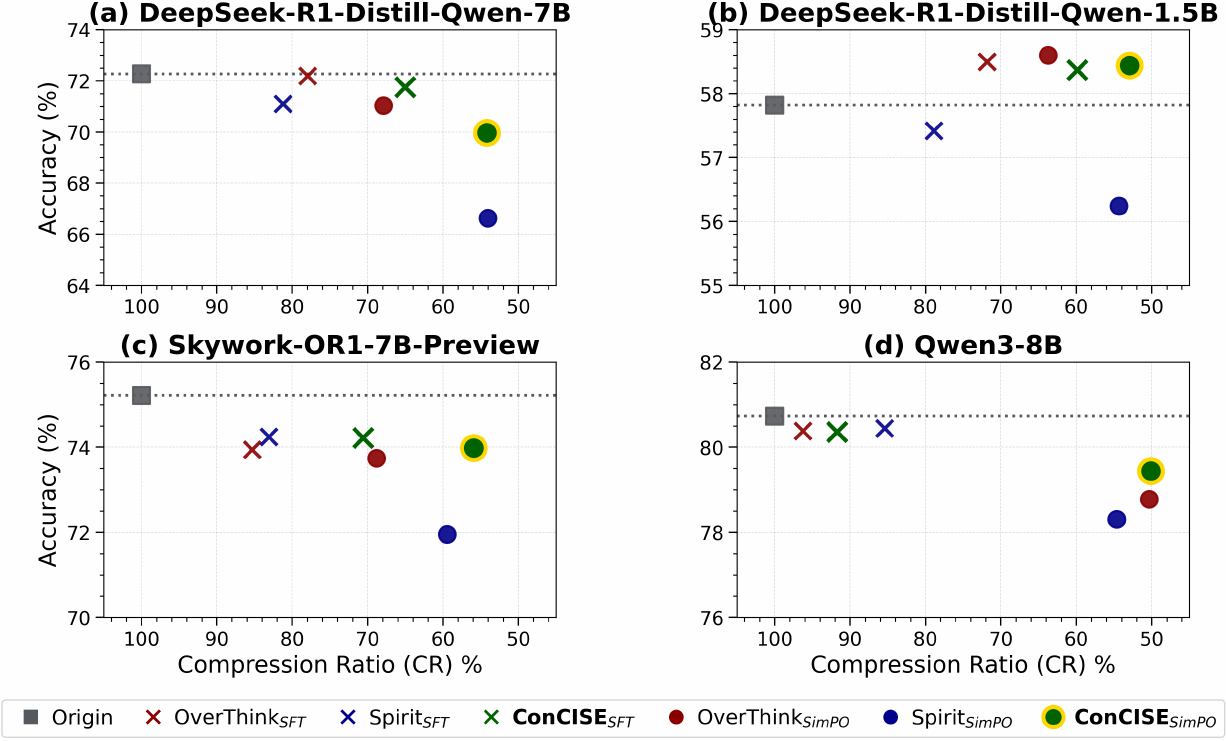}
\caption{\textsc{ConCISE} achieves a better trade-off between compression and task performance than baselines.}
\label{fig:Performance}
\end{figure}

To mitigate LRM output verbosity, recent research focuses on compressing their reasoning chains \cite{qu2025survey, sui2025stop}. A prominent strategy involves fine-tuning LRMs on concise reasoning datasets, enabling them to generate shorter responses \cite{ma2025cot, chen2024not, munkhbat2025self}. The effectiveness of this compression strategy hinges on the training dataset design. As shown in Figure~\ref{fig:Comparison}, common sampling-based selection, which generates multiple candidates and picks the shortest correct one, or further removes post-answer redundant reflections \cite{team2025kimi, chen2024not}, lacks control during generation, potentially leaving unnecessary steps and reducing compression effectiveness. Another approach, post-hoc pruning, identifies and removes redundant or less important steps from reasoning chains \cite{cui2025stepwise, xia2025tokenskip}, risks disrupting reasoning coherence and degrading the performance of LRMs after fine-tuning.

To overcome existing limitations, we aim to construct compact, coherent reasoning chains as training datasets by precisely removing redundant reflections, ensuring LRMs do not suffer performance degradation after fine-tuning. To this end, based on the understanding that reflections are not solely determined by correctness \cite{yang2025dynamic}, we thus adopt a \textbf{confidence-guided} perspective to understand the generation of reflection steps in LRM's reasoning processes. This perspective offers a clear interpretation of two key patterns of redundancy: \textit{Confidence Deficit}, where low internal confidence causes models to undertrust and reflect on correct intermediate steps; and \textit{Termination Delay}, where reflection persists despite a repeatedly verified answer. These patterns inflate reasoning chains and provide actionable insights for how to create concise reasoning chains.

Therefore, we propose \textsc{ConCISE} (\textbf{Con}fidence-guided \textbf{C}ompression \textbf{I}n \textbf{S}tep-by-step \textbf{E}fficient reasoning), a framework that leverages a confidence-guided perspective for constructing concise reasoning data by actively suppressing redundant reflection during generation. \textsc{ConCISE} features two complementary components designed to mitigate \textit{Confidence Deficit} and \textit{Termination Delay}, respectively: \textit{Confidence Injection}, which inserts tailored phrases before potential reflection points to boost the model's internal confidence, consequently curbing unnecessary reflection; and \textit{Early Stopping}, which employs a lightweight confidence detector to monitor internal confidence after an answer is reached, terminating reasoning upon achieving sufficient confidence. The synergy of these mechanisms enables \textsc{ConCISE} to produce more efficient and compact reasoning chains. 

We then fine-tune four mainstream LRMs using \textsc{ConCISE}-generated data via SFT \cite{wei2021finetuned} and SimPO \cite{meng2024simpo}, evaluating their performance across multiple reasoning benchmarks. As shown in Figure~\ref{fig:Performance}, Experimental results demonstrate that \textsc{ConCISE} consistently achieves a superior trade-off between reasoning compression and task performance under both SFT
and SimPO settings compared to baselines. Notably, \textsc{ConCISE} reduces average response length by approximately 50\% under SimPO while maintaining high accuracy. Our subsequent analysis reveals that \textsc{ConCISE}'s unique training data design enables LRMs, after fine-tuning, to learn to strategically avoid generating redundant reflection steps without harming essential critical reasoning content, thereby achieving efficient reasoning compression while task performance is well maintained.


\section{Related Work}
\label{sec:related_work}

\begin{figure*}[t]
\centering
\includegraphics[width=\linewidth]{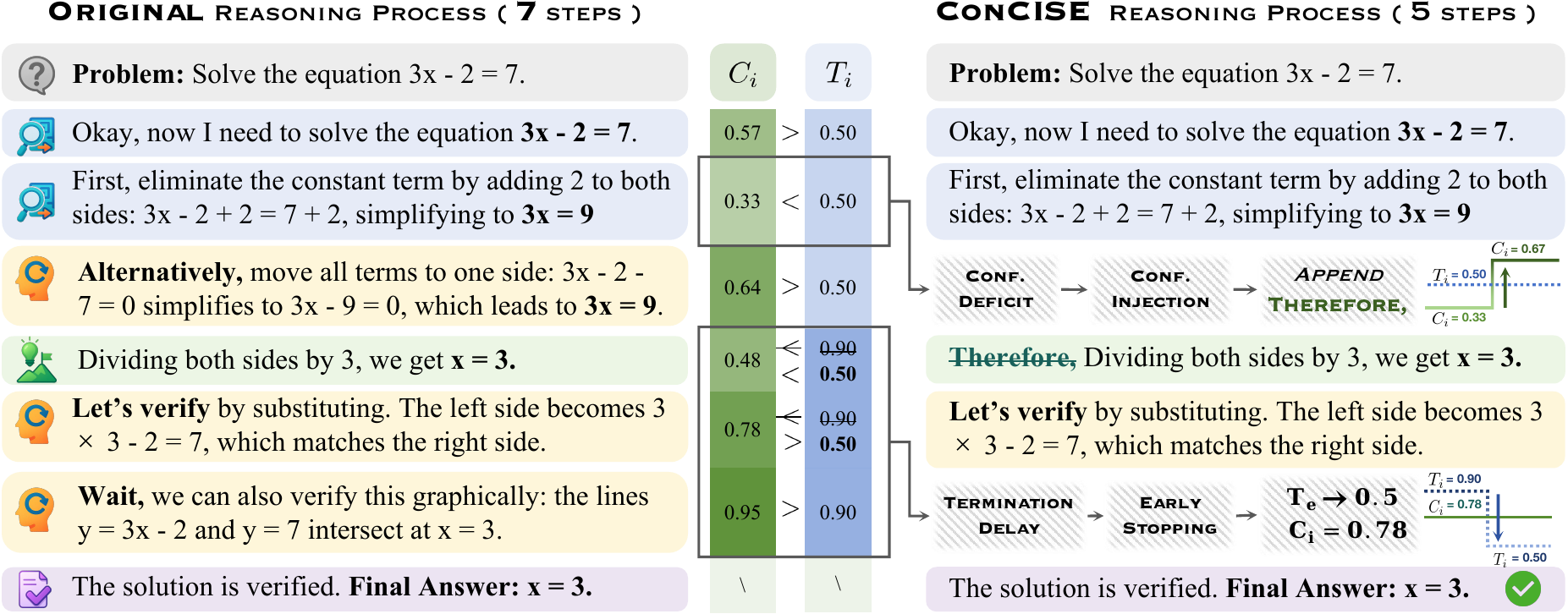}
\caption{Illustration of \textsc{ConCISE}'s confidence-guided approach: identifying patterns (\textit{Confidence Deficit}, \textit{Termination Delay}) and applying mechanisms (\textit{Confidence Injection}, \textit{Early Stopping}) to suppress redundant reflections, shown in contrast to the original reasoning process. \(C_i\) denotes step confidence and \(T_i\) its threshold.}
\label{fig:example}
\end{figure*}

Recent research has increasingly focused on mitigating verbosity and redundancy in reasoning chains generated by LRMs, which often produces long outputs that increase computational costs, and even degrade accuracy \cite{wu2025more, nayab2024concise, wang2025thoughts}. To address this problem, existing approaches can be broadly categorized into three classes: Input-based, Decoding-based, and Model-based \cite{sui2025stop}.

\noindent\textbf{Input-based Methods} aim to promote concise reasoning by modifying input texts \cite{lee2025well}. Common techniques include imposing token limits in prompts or instructing the model to reason briefly \cite{han2024token, renze2024benefits}. However, these approaches often fail when the model does not consistently follow these instructions. Another approach, task routing, dynamically selects an appropriate model or reasoning strategy based on input characteristics like question complexity \cite{ong2024routellm, chuang2025confident, aytes2025sketch}. Its effectiveness, however, depends on the accurate assessment of input characteristics, which is not always reliable.

\noindent\textbf{Decoding-based Methods} intervene during decoding to control the reasoning process. One common approach compresses steps into latent representations rather than explicit text, which improves brevity but sacrifices interpretability \cite{hao2024training, shen2025codi}. Another strategy uses dynamic decoding, evaluating each reasoning step and deciding whether to modify or discard it during generation \cite{sun2024fast, zhang2025lightthinker, xu2025chain}. While effective at reducing verbosity, these methods introduce additional computational overhead and may disrupt the reasoning coherence.

\noindent\textbf{Model-based Methods} train models to generate concise reasoning directly. One approach employs reinforcement learning with reward functions that penalize verbosity \cite{shen2025dast, aggarwal2025l1, luo2025o1, arora2025training, yu2025dapo}, but can be sensitive to reward formulation and impose significant costs. Another common approach is to fine-tune LRMs on datasets with concise reasoning chains\cite{ma2025cot, cui2025stepwise, xia2025tokenskip}. These datasets are often created by selecting the shortest correct chain or removing redundant parts post-hoc. Such methods may inadvertently retain redundant reflection steps or remove useful context, degrading compression efficiency or model performance.

Motivated by the limitations of existing methods, we propose \textsc{ConCISE}, a model-based method that precisely identifies and actively suppresses redundant reflection steps throughout the reasoning process. This approach enables the model to generate efficient reasoning chains while preserving reasoning coherence. Leveraging \textsc{ConCISE}, we construct a high-quality training dataset and fine-tune the model to adopt this concise reasoning ability without compromising model performance.
\section{Method}

\subsection{Confidence-guided Formulation}
Reflections in \textsc{LRMs} are not solely triggered by correctness; in many cases, reflection steps are invoked even on correct steps that have been verified \cite{yang2025dynamic}. This suggests that reflection behavior is also linked to the model’s internal confidence about current reasoning rather than correctness alone. To explain this, we adopt a \textbf{confidence-guided} perspective to formalize when and why \textsc{LRMs} engage in reflection.

Let $S_i = \{s_1, s_2, \ldots, s_i\}$ denote the partial reasoning chain up to step $i$, where each $s_i$ is a textual reasoning unit. We associate each step $s_i$ with a confidence score $c_i \in [0,1]$, representing the model’s internal confidence of that step. The generation policy of the LRM, denoted by $\pi_\theta$, maps the current reasoning context $S_i$ to the next step $s_{i+1}$.

To model the decision between proceeding and reflecting, we introduce a dynamic threshold $t_i \in [0,1]$, which may vary with the model or the context. At each step, the model generates a \texttt{ReflectionStep} as $s_{i+1}$ if its current internal confidence $c_i$ falls below the threshold $t_i$ (i.e., $c_i < t_i$). From this perspective, we further analyze the reflection behavior of \textsc{LRMs} and formally articulate two key patterns responsible for reflection-related redundancy existing in the reasoning process: \textit{Confidence Deficit} and \textit{Termination Delay}.

\subsection{Two Key Patterns of Redundancy}

\paragraph{Confidence Deficit.}  
One major source of redundancy in LRMs stems from their tendency to undertrust their correct intermediate steps. LRMs often display unexpected reflection despite exhibiting fine-grained reasoning capabilities and achieving high stepwise accuracy, triggering reflection even on simple and unambiguous reasoning steps. This self-undermining behavior leads to redundant reflection with minimal semantic gain. We refer to this phenomenon as \textit{Confidence Deficit}. Formally, \textit{Confidence Deficit} can be defined as the occurrence of a step $s_i$ satisfying:
\begin{equation*} 
\exists i , \quad c_i < t_i \quad \text{despite } s_i \text{ being correct}.
\end{equation*}

\paragraph{Termination Delay.}  
LRMs exhibit another important redundant reasoning pattern we term \textit{Termination Delay}: After producing a confident final answer, the model is expected to conclude with minimal additional reasoning. However, it often continues to generate unnecessary reflection steps even after repeatedly verifying the same conclusion. This behavior can be attributed to a sharp increase in the confidence threshold $t_i$ after reaching the answer at step $i'$, making it increasingly difficult for the model's internal confidence $c_i$ to exceed $t_i$. As a result, even when $c_i$ becomes relatively high, it may still fall short of the heightened $t_i$, leading to verbose post-answer reasoning. Formally, \textit{Termination Delay} can be characterized as:
\begin{equation*} 
\exists i > i', \; \forall j < i', \quad t_i \gg t_j \; \text{and} \; c_i < t_i
\end{equation*}

\subsection{The \textsc{ConCISE} Framework}
To mitigate the above patterns—\textit{Confidence Deficit} and \textit{Termination Delay}—we propose \textsc{ConCISE}, a framework that dynamically steers the reasoning generation process to reduce redundant reflections and produce concise reasoning chains. \textsc{ConCISE} integrates two mechanisms: \textit{Confidence Injection}, which actively inserts confidence phrases to suppress unnecessary reflection steps, and \textit{Early Stopping}, which halts generation once sufficient post-answer confidence is detected. These two mechanisms operate together to generate concise reasoning chains without compromising inference quality. The overall process is illustrated in Figure~\ref{fig:example}.

\subsubsection{Confidence Injection}
To alleviate \textit{Confidence Deficit}, we design a simple yet effective mechanism called \textit{Confidence Injection}. The key idea is to boost the model's internal confidence during reasoning by inserting designed phrases, referred to as \textit{confidence phrases}, to prevent unnecessary reflection steps. Considering that indiscriminate injection could disrupt reasoning, we selectively insert confidence phrases at critical points where the model's confidence is relatively low, that is, when $c_i < t_i$, a condition that signifies an impending reflection step. 

Specifically, at each step $s_i$, the model first generates $s_{i+1} = \pi_\theta(S_i)$. If $s_{i+1}$ is a reflection step, we retroactively modify the input $S_i$ by appending a confidence phrase $p_i$ sampled from a curated pool $\mathcal{P}$, and regenerate $s_{i+1}$. Formally, the updated generation process is defined as:
\begin{equation*} 
s_{i+1} = \pi_\theta(S_i) \quad \text{if} \; c_i \geq t_i \; \text{else} \; \pi_\theta(S_i , p_i)
\end{equation*}

We manually constructed the initial version of $\mathcal{P}$. To evaluate and further refine the initial pool, we conducted experiments measuring the effectiveness of different phrases. We inserted each candidate phrase before the reflection steps and recorded the probability that the model still generated a reflection step. This probability indicates whether the updated confidence $c_i^*$ (after phrase insertion) exceeds the dynamic threshold $t_i$. As shown in Figure~\ref{fig:refprob}, different phrases exhibit varying reflection rates. Notably, even the best-performing phrases still yield a reflection rate around 20\%, suggesting that confidence injection reduces redundant reflections while still retaining necessary verifications. We then selected 20 phrases with the lowest reflection rates to form the final confidence phrase pool. The composition of the phrase pool and detection of reflection steps are provided in Appendix~\ref{cha:conf_pool}.

\subsubsection{Early Stopping}

\begin{figure}[t]
\centering

\begin{subfigure}[t]{0.60\linewidth}
    \centering
    \includegraphics[width=\linewidth]{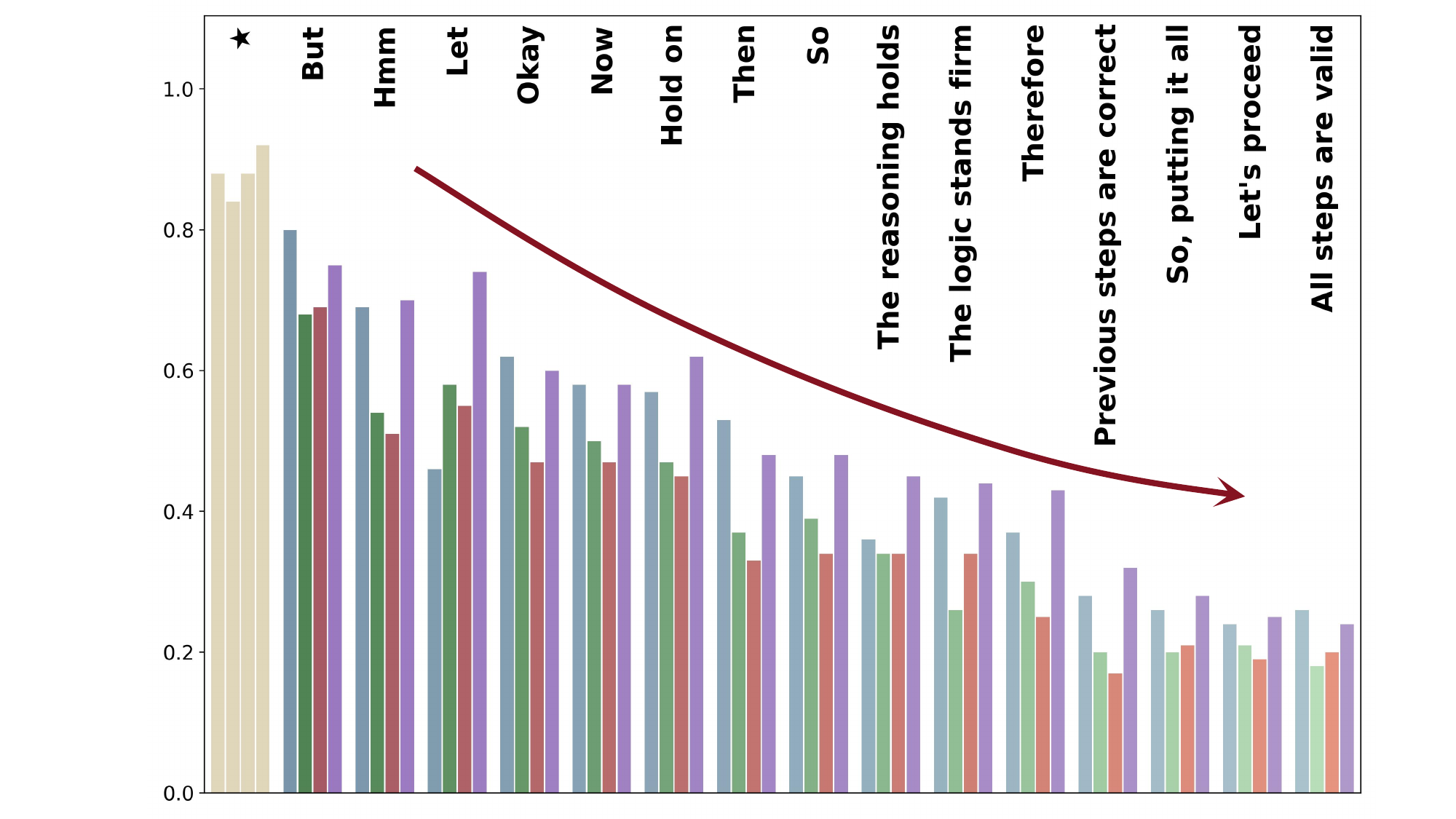}
    \caption{Reflection probability of next step after phrase injection.}
    \label{fig:refprob}
\end{subfigure}
\hfill
\begin{subfigure}[t]{0.38\linewidth}
    \centering
    \includegraphics[width=\linewidth]{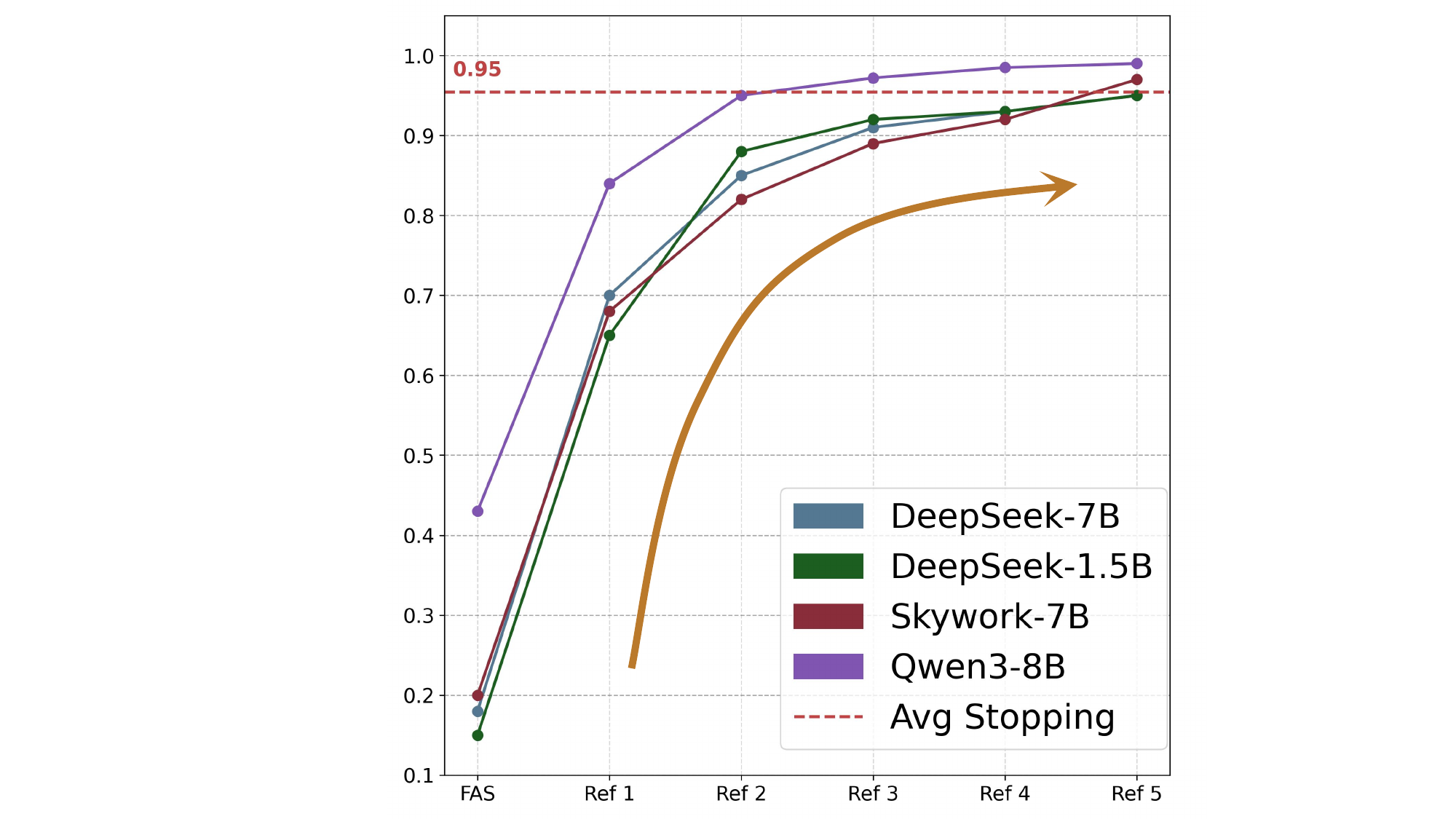}
    \caption{Confidence calculated by our detector.}
    \label{fig:confidence}
\end{subfigure}

\caption{Effectiveness and necessity of Confidence Injection and Termination Delay(details in Appendix~\ref{cha:datasets_details}).}
\label{fig:verify}
\end{figure}

Although Confidence Injection effectively boosts the model's confidence during reasoning, it does not fully mitigate \textit{Termination Delay}, which is caused by the high threshold $t_i$ after the model reaches an answer, leading the model to continue unnecessary reflections even when its confidence is already high. To address this, we design an \textit{Early Stopping} mechanism based on direct estimation of the model's internal confidence. 

Specifically, we construct a lightweight confidence detector to provide a quantitative proxy for the model's internal confidence about the answer. We introduce a probing prompt after the current reasoning context and analyze the probability distribution of generated continuations. Based on the statistical analysis shown in Appendix~\ref{cha:early_prompt}, we collect a set of confidence-indicative phrases $\mathcal{W}_{+}$, which reflect affirmations of high certainty. The detected confidence score $\hat{c}_i$ at step $i$ is calculated as:
\begin{equation*} 
\hat{c}_i = \sum_{w \in \mathcal{W}_{+}} p(w \mid S_i, \text{Probing prompt}),
\end{equation*}

where $p(\cdot|\cdot)$ denotes the continuation probabilities determined by $\pi_\theta$. We then calculate the average $\hat{c}_i$ after the First Answer Step (FAS) is generated, including the subsequent five reflection steps (Ref1-5). As shown in Figure~\ref{fig:confidence}, the model exhibits relatively low confidence at the FAS, but increases sharply after the first reflection, and continues to rise through subsequent reflections. Notably, when the reasoning chain stops, the average confidence is $0.95$, motivating the introduction of a manually controlled lower threshold $t_e$. We ultimately set $t_e = 0.5$ through experiments.

Additionally, the application of the threshold is carefully controlled to prevent premature termination: once the detected confidence $\hat{c}_i$ exceeds $t_e$, the model is first prompted to output a final answer (e.g., by appending \texttt{Final Answer:}). The reasoning process halts only if this answer is subsequently verified as correct; otherwise, generation continues. More details regarding the selection of $t_e$ and \textit{Early Stopping} are provided in Appendix~\ref{cha:early_prompt}.

\subsection{Building Efficient Reasoning Chains}
{
\footnotesize
\begin{spacing}{0.8}
\begin{algorithm}[t] 
\caption{Workflow of \textsc{ConCISE}.}
\label{alg: concise} 
\textbf{Preparation:} 
LRM generation policy: $\pi_\theta$,  Confidence Phrase Pool: $\mathcal{P}$, Early Stopping Threshold $t_e$, Prompt Template $T$\;

\KwIn{Question $q$, Ground Truth $gt$} 

\textbf{Initialize:} 
Reasoning chain $S_0 \leftarrow T(q)$\; 

\For{each reasoning step $i = 1, 2, \ldots$}{ 
    $s_i \leftarrow \pi_\theta(S_{i-1})$\;
    \If{$s_i$ is a reflection step}{ 
        Sample a confidence phrase $p_i \in \mathcal{P}$
        $s_i \leftarrow \pi_\theta(S_{i-1}, p_i)$
    }
    $S_i \leftarrow S_{i-1} + s_i$\;
    Compute detected confidence $\hat{c}_i$\;
    \If{$\hat{c}_i > t_e $}{
        $a \leftarrow \pi_\theta(S_{i}, \text{'Final Answer:'})$\;
        \If{$\text{isequal}(a,gt)$}{
            \textbf{break}\; 
        }
    }
} 
\If{$\text{isequal}(a,gt)$}{
    Summary: $S \leftarrow S_i + \pi_\theta(S_i, '\text{</think>}')$\;
    \KwOut{\textsc{ConCISE} reasoning chain $S$} 
}
\Else{ 
    $S_i$ and $q$ are \textbf{discarded}
}
\end{algorithm}
\end{spacing}
}
\begin{table*}[t]
\centering
\small
\renewcommand{\arraystretch}{1.0}
\setlength{\heavyrulewidth}{1.2pt} 
\setlength{\lightrulewidth}{0.8pt}  
\resizebox{\textwidth}{!}{
\begin{tabular}{l|ccc|ccc|ccc|ccc|cc}
\toprule
\textbf{Model} &
\multicolumn{3}{c|}{\textbf{Math-500}} &
\multicolumn{3}{c|}{\textbf{GSM8K}} &
\multicolumn{3}{c|}{\textbf{AIME24}} &
\multicolumn{3}{c|}{\textbf{GPQA}} &
\multicolumn{2}{c}{\textbf{Average}} \\
\cmidrule(r){2-4} \cmidrule(r){5-7} \cmidrule(r){8-10} \cmidrule(r){11-13} \cmidrule(r){14-15}
& Acc.$\uparrow$ & Tok.$\downarrow$ & CR$\downarrow$ & Acc.$\uparrow$ & Tok.$\downarrow$ & CR$\downarrow$ & Acc.$\uparrow$ & Tok.$\downarrow$ & CR$\downarrow$ & Acc.$\uparrow$ & Tok.$\downarrow$ & CR$\downarrow$ & Acc.$\uparrow$ & CR$\downarrow$ \\
\midrule
\rowcolor{gray!15} 
\textbf{DeepSeek-7B}\textsubscript{\textcolor[rgb]{0.5,0.0,0.0}{Origin}}                           & 90.8 & 3854 & 100\% & 93.1 & 1442 & 100\% & 54.2 & 13574 & 100\% &51.0 &8142 &100\% & 72.3 & 100\%\\
\cmidrule{1-15}
OverThink\textsubscript{\textcolor[rgb]{0.0,0.5,0.0}{SFT}} & 92.2 & 2538 & 66\% & 93.0 & 1002 & 70\% & 52.5 & 11225 & 83\% & 51.1 & 7639 & 94\% & \textbf{72.2} & \underline{78\%}\\
Spirit\textsubscript{\textcolor[rgb]{0.0,0.5,0.0}{SFT}}    & 91.0 & 2935 & 76\% & 91.4 & 1107 & 77\% & 51.7 & 11529 & 85\% & 50.3 & 7084 & 87\% & 71.1 & 81\%\\
\textbf{ConCISE}\textsubscript{\textcolor[rgb]{0.0,0.5,0.0}{SFT}}   & 92.0 & 2244 & 58\% & 92.9 & 832  & 58\% & 52.1 & 9751 & 72\% &50.0 & 5892 & 72\% & \underline{71.8} & \textbf{65}\%\\
\cmidrule{1-15}
OverThink\textsubscript{\textcolor[rgb]{0.0,0.2,0.6}{SimPO}} & 91.4 & 2405 & 62\% & 92.9 & 879  & 61\% & 50.0 & 9603  & 71\% & 49.9 & 6305 & 77\% &\textbf{71.0} &68\% \\
Spirit\textsubscript{\textcolor[rgb]{0.0,0.2,0.6}{SimPO}}    & 87.2 & 1765 & 46\% & 90.8 & 688  & 48\% & 38.3 & 6926  & 51\% & 50.2 & 5832 & 72\% &66.6 &\textbf{54\%}\\
\textbf{ConCISE}\textsubscript{\textcolor[rgb]{0.0,0.2,0.6}{SimPO}}   & 91.0 & 1946 & 51\% & 92.1 & 715  & 50\% & 48.3 & 7745  & 57\% & 48.0 & 4859 & 60\% &\underline{70.0} &\textbf{54\%}\\

\specialrule{1.2pt}{2.5pt}{2.5pt}

\rowcolor{gray!15} 
\textbf{DeepSeek-1.5B}\textsubscript{\textcolor[rgb]{0.5,0.0,0.0}{Origin}}                         & 82.2 & 4784 & 100\% & 85.4 & 2219 & 100\% & 29.2 & 17465 &100\% & 34.5 & 9492 &100\% & 57.8 & 100\%\\
\cmidrule{1-15}
OverThink\textsubscript{\textcolor[rgb]{0.0,0.5,0.0}{SFT}} & 84.0 & 3296 & 69\% & 85.2 & 1200 & 54\% & 30.0 & 12893 & 74\% & 34.8 & 8600 & 91\% & \textbf{58.5} & \underline{72\%}\\
Spirit\textsubscript{\textcolor[rgb]{0.0,0.5,0.0}{SFT}}    & 83.8 & 3857 & 81\% & 84.8 & 1476 & 67\% & 28.8 & 13685 &78\% & 32.3 & 8555 & 90\% & 57.4 & 79\%\\
\textbf{ConCISE}\textsubscript{\textcolor[rgb]{0.0,0.5,0.0}{SFT}}   & 83.6 & 2701 & 57\% & 84.9 & 923  & 42\% & 30.0 & 11359 & 65\% &35.0 & 7253 & 76\% & \underline{58.4} & \textbf{60\%}\\
\cmidrule{1-15}
OverThink\textsubscript{\textcolor[rgb]{0.0,0.2,0.6}{SimPO}} & 83.6 & 2738 & 57\% & 84.8 & 1003 & 45\% & 31.3 & 11465 & 66\% &34.7 & 8250 &87\% & \textbf{58.6} & 64\%\\
Spirit\textsubscript{\textcolor[rgb]{0.0,0.2,0.6}{SimPO}}    & 82.6 & 2455 & 51\% & 82.6 & 804  & 36\% & 26.7 & 9946  & 57\% &33.1 &6910 &73\% & 56.2 & \underline{54\%}\\
\textbf{ConCISE}\textsubscript{\textcolor[rgb]{0.0,0.2,0.6}{SimPO}}   & 83.6 & 2429 & 51\% & 84.3 & 803  & 36\% & 30.4 & 8810  & 50\% & 35.4 & 7056 &74\% & \underline{58.4} & \textbf{53}\%\\

\specialrule{1.2pt}{2.5pt}{2.5pt}

\rowcolor{gray!15} 
\textbf{Skywork-7B}\textsubscript{\textcolor[rgb]{0.5,0.0,0.0}{Origin}}                            & 93.6 & 4178 & 100\% & 93.2 & 2111 & 100\% & 62.9 & 12464 & 100\% & 51.1 & 8374 & 100\% & 75.2 & 100\% \\
\cmidrule{1-15}
OverThink\textsubscript{\textcolor[rgb]{0.0,0.5,0.0}{SFT}} & 92.8 & 3410 & 82\% & 93.1 & 1561 & 74\% & 59.2 & 11047 & 89\% & 50.7 & 8128 & 97\% & 73.9 & 85\% \\
Spirit\textsubscript{\textcolor[rgb]{0.0,0.5,0.0}{SFT}}    & 93.4 & 3279 & 79\% & 93.1 & 1452 & 69\% & 58.8 & 11815 & 95\% & 52.0 & 7565 & 90\% & \textbf{74.2} & \underline{83\%} \\
\textbf{ConCISE}\textsubscript{\textcolor[rgb]{0.0,0.5,0.0}{SFT}}   & 93.2 & 2740 & 66\% & 92.7 & 1247 & 59\% & 59.2 & 9871 & 79\% & 51.8 & 6543 & 78\% & \textbf{74.2} & \textbf{71}\% \\
\cmidrule{1-15}
OverThink\textsubscript{\textcolor[rgb]{0.0,0.2,0.6}{SimPO}} & 93.8 & 2758 & 66\% & 93.4 & 1114 & 53\% & 56.3 & 9422 & 76\% & 51.5 & 6773 & 81\% & \underline{73.7} & 69\% \\
Spirit\textsubscript{\textcolor[rgb]{0.0,0.2,0.6}{SimPO}}    & 92.0 & 2329 & 56\% & 92.3 & 801 & 38\% & 52.5 & 8055 & 65\% & 51.0 & 6663 & 80\% & 72.0 & \underline{59\%} \\
\textbf{ConCISE}\textsubscript{\textcolor[rgb]{0.0,0.2,0.6}{SimPO}}   & 93.0 & 2207 & 53\% & 93.3 & 882 & 42\% & 55.9 & 7598 & 61\% & 51.7 & 5668 & 68\% & \textbf{74.0} & \textbf{56}\% \\

\specialrule{1.2pt}{2.5pt}{2.5pt}

\rowcolor{gray!15} 
\textbf{Qwen3-8B}\textsubscript{\textcolor[rgb]{0.5,0.0,0.0}{Origin}}                               & 93.4 & 5142 & 100\% & 95.6 & 2211 & 100\% & 73.3 & 15094 & 100\% & 60.6 & 7309 & 100\% & 80.7 & 100\% \\
\cmidrule{1-15}
OverThink\textsubscript{\textcolor[rgb]{0.0,0.5,0.0}{SFT}} & 93.2 & 4963 & 97\% & 95.5 & 2133 & 96\% & 72.5 & 14123 & 94\% & 60.3 & 7204 & 99\% & \underline{80.4} & 96\% \\
Spirit\textsubscript{\textcolor[rgb]{0.0,0.5,0.0}{SFT}}    & 93.2 & 4184 & 81\% & 95.2 & 1675 & 76\% & 73.3 & 13854 & 92\% & 60.0 & 6784 & 93\% & \underline{80.4} & \textbf{85\%} \\
\textbf{ConCISE}\textsubscript{\textcolor[rgb]{0.0,0.5,0.0}{SFT}}   & 93.0 & 4712 & 92\% & 95.6 & 1907 & 86\% & 72.1 & 14168 & 94\% & 60.4 & 6952 & 95\% & \textbf{80.6} & \underline{92\%} \\
\cmidrule{1-15}
OverThink\textsubscript{\textcolor[rgb]{0.0,0.2,0.6}{SimPO}} & 93.6 & 2488 & 48\% & 95.2 & 929 & 42\% & 69.2 & 9468 & 63\% & 57.1 & 3524 & 48\% & \underline{78.9} & \textbf{50\%} \\
Spirit\textsubscript{\textcolor[rgb]{0.0,0.2,0.6}{SimPO}}    & 93.2 & 2920 & 57\% & 95.3 & 997 & 45\% & 67.9 & 10441 & 69\% & 56.8 & 3473 & 48\% & 78.3 & 55\% \\
\textbf{ConCISE}\textsubscript{\textcolor[rgb]{0.0,0.2,0.6}{SimPO}}   & 93.0 & 2271 & 44\% & 95.5 & 841 & 38\% & 71.6 & 10098 & 67\% & 57.6 & 3751 & 51\% & \textbf{79.4} & \textbf{50}\% \\

\bottomrule
\end{tabular}}
\caption{Comparison of OverThink, SPIRIT, and ConCISE methods, fine-tuned with SFT or SimPO on four LRMs. Metrics include Accuracy (Acc.), Token Count (Tok.), and Compression Ratio (CR) over four benchmarks. \textcolor[rgb]{0.25,0.25,0.25}{Light gray rows} denote original LRM performance. Best and second-best results are in \textbf{bold} and \underline{underlined} respectively. Note that the ``Average'' column is computed as an unweighted mean across benchmarks, and this convention applies to all subsequent tables as well.}
\label{tab:all_eval}
\end{table*}

\textsc{ConCISE} integrates \textit{Confidence Injection} and \textit{Early Stopping} to dynamically control the reasoning generation process. Given an input question, the \textsc{LRM} generates the reasoning chain step-by-step. At each step $s_i$: if it is identified as a reflection step, \textit{Confidence Injection} is applied by appending a confidence prompt and regenerating $s_i$ with boosted internal confidence. $s_i$ is then added to the current chain $S_i$. Subsequently, a lightweight confidence detector estimates the detected confidence score $\hat{c}_i$. If $\hat{c}_i>t_e$, the \textsc{LRM} is prompted to provide a final answer $a$. Crucially, the generation process terminates early at this stage if, and only if, this answer $a$ matches the ground truth $gt$ and $\hat{c}_i>t_e$, otherwise reasoning continues to the next step. After the loop concludes, should a correct final answer $a$ have been verified, a concluding summary (e.g., prompted by \texttt{</think>}) is generated and appended to the accumulated steps $S_i$ to form the complete \textsc{ConCISE} reasoning chain $S$, otherwise, if a wrong answer was finally achieved, the partial chain and the original question are discarded. Algorithm~\ref{alg: concise} summarizes this coordinated reasoning process in \textsc{ConCISE}.

\subsection{Training Objective and Fine-tuning}
To align LRM generation with the behavior encouraged by \textsc{ConCISE}, we first construct a fine-tuning dataset by applying the \textsc{ConCISE} pipeline to a set of questions. This yields concise reasoning chains in which redundant reflections are actively suppressed. We then fine-tune the LRM on this dataset using two standard learning paradigms: Supervised Fine-Tuning (SFT) and Simple Preference Optimization (SimPO). Both approaches optimize a shared objective that governs the desired generation behavior:
\begin{equation*} 
\begin{cases} 
\pi_{\theta}(S_i) \rightarrow \pi_{\theta}(S_i + p_i), & \text{if } c_i < t_i \\
\pi_{\theta}(S_i) \rightarrow \text{Terminate}, & \text{if } i \geq i' \text{ and } \hat{c}_i > t_e
\end{cases} 
\end{equation*}

Specifically, when the model's confidence \( c_i \) at a given step falls below the threshold \( t_i \), it learns to approximate the distribution conditioned on an injected confidence phrase \( p_i \), which helps improve internal certainty and suppress unnecessary reflections. After the first answer is generated, if the detected confidence score \( \hat{c}_i \) exceeds the early stopping threshold \( t_e \), the model is explicitly trained to terminate reasoning at that point. Through fine-tuning, the model could learn to regulate its generation trajectory based on confidence, strengthening certainty during reasoning and halting once sufficient confidence is reached.

\begin{figure*}[t]
\centering

\begin{subfigure}[t]{0.32\linewidth}
    \centering
    \includegraphics[width=\linewidth]{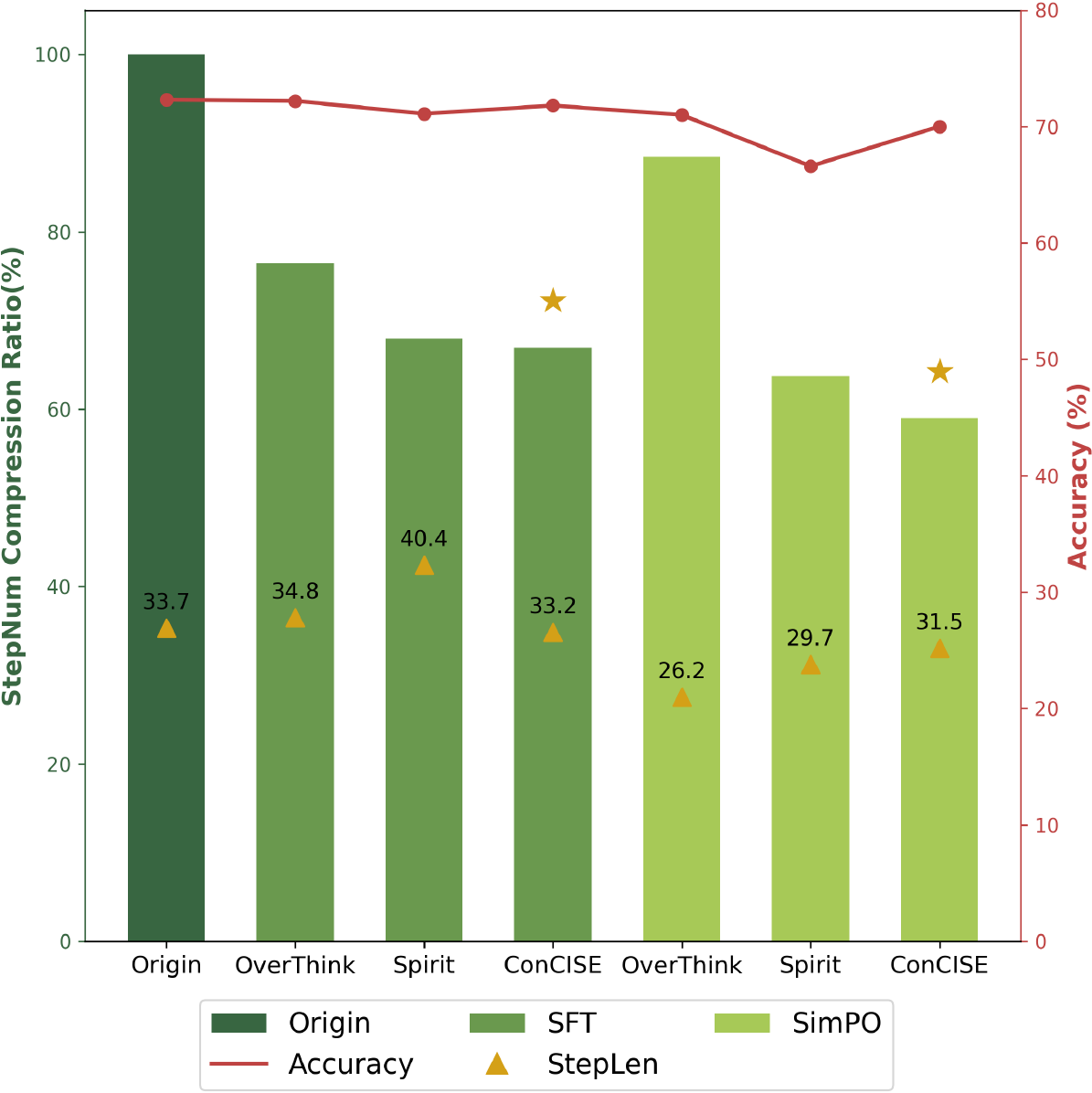}
    \caption{Average \texttt{Acc}, \texttt{StepNum}, and \texttt{StepLen}.}
    \label{fig:steplen}
\end{subfigure}
\hfill
\begin{subfigure}[t]{0.3\linewidth}
    \centering
    \includegraphics[width=\linewidth]{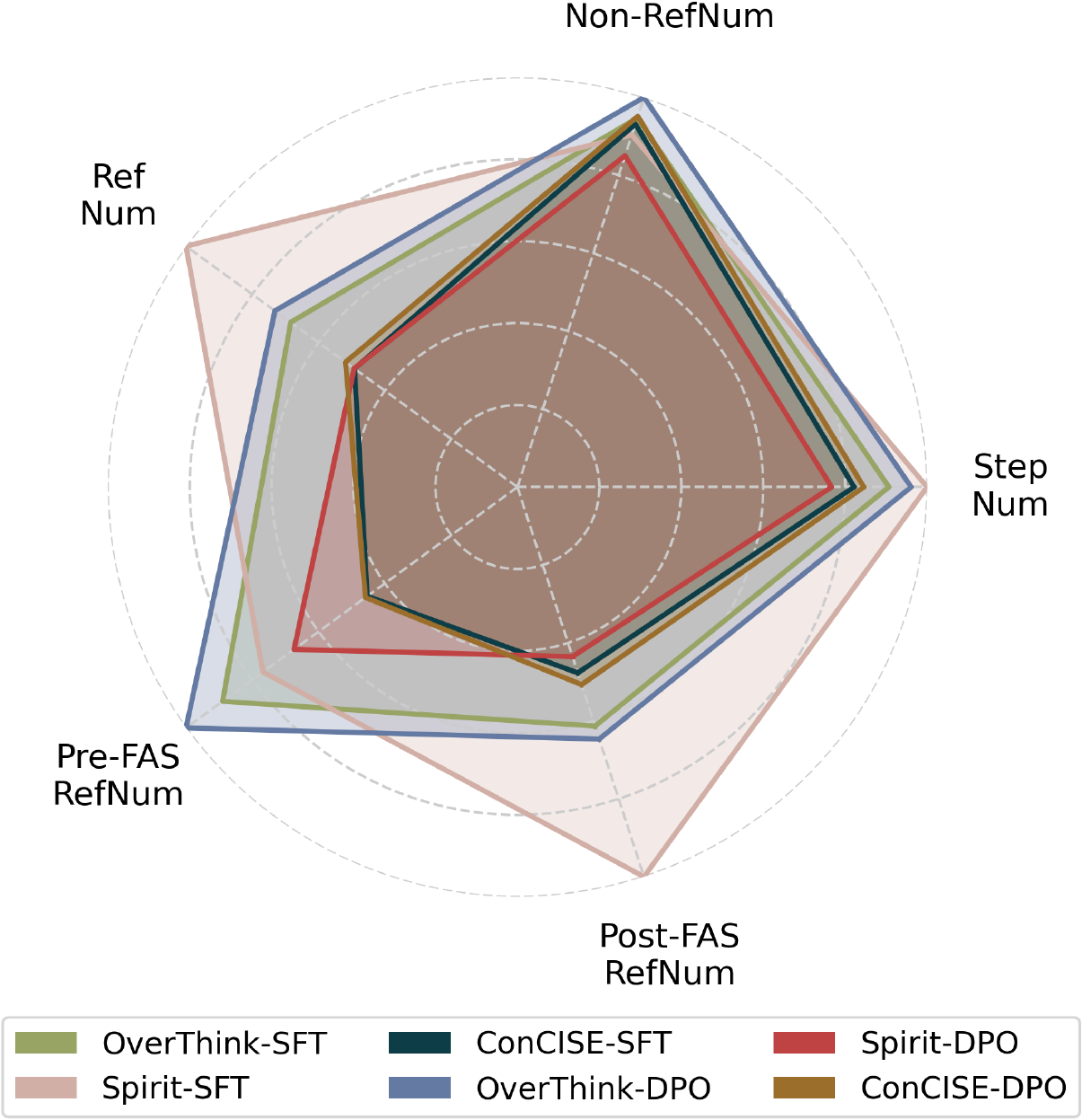}
    \caption{Analysis of Reflection steps.}
    \label{fig:math_metric}
\end{subfigure}
\hfill
\begin{subfigure}[t]{0.3\linewidth}
    \centering
    \includegraphics[width=\linewidth]{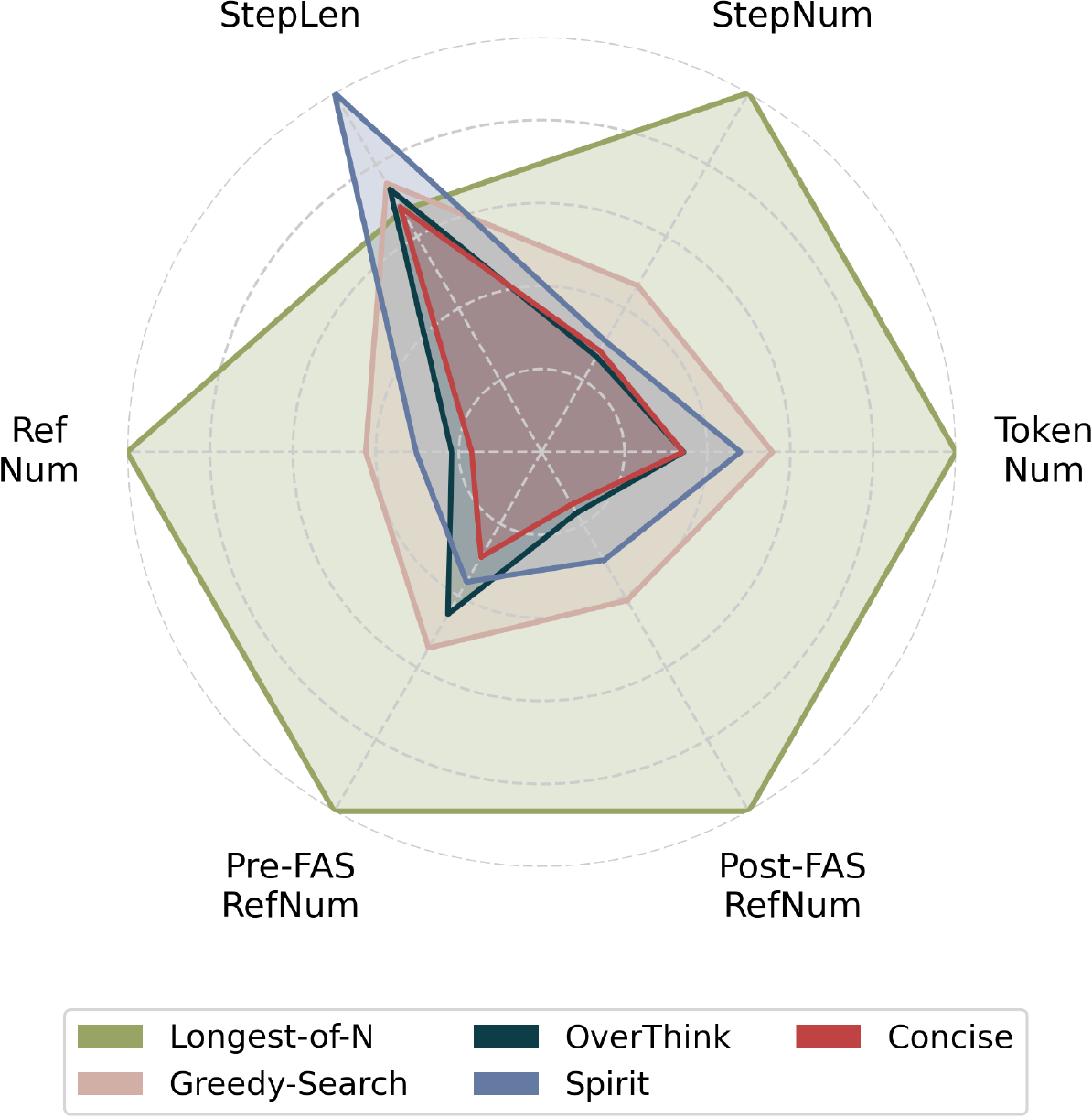}
    \caption{Metrics of training datasets.}
    \label{fig:train_metric}
\end{subfigure}

\caption{Further analysis of reasoning chains and training datasets on DeepSeek-R1-Distill-Qwen-7B.}
\label{fig:analysis}
\vspace{-2ex}
\end{figure*}

\section{Experiments}
\label{sec:exp}

\subsection{Settings}

\paragraph{Models and Datasets.}  
We evaluate ConCISE on four LRMs: DeepSeek-R1-Distill-Qwen-7B \& 1.5B\cite{guo2025deepseek}, Skywork-OR1-7B-Preview\cite{skywork-or1-2025}, and Qwen3-8B\cite{qwen3}. As for the training dataset, we select 2,000 questions from the MATH training set\cite{hendrycksmath2021}, ensuring that each question yields a correct answer under greedy decoding and maintaining diversity in difficulty and response lengths. Importantly, for each model, we generate its own fine-tuning data using the \textsc{ConCISE} pipeline, rather than relying on external models, which ensures consistency between the training data and the reasoning style of the target model.

\paragraph{Evaluation.}  
\label{sec:analysis}
We evaluate model performance across four benchmarks: GSM8K\cite{cobbe2021training}, Math-500\cite{hendrycksmath2021}, AIME24, and GPQA\_diamond\cite{rein2024gpqa}. All evaluations use the same decoding configuration with temperature = 0.6 and top\_p = 0.95\cite{guo2025deepseek}. The maximum length is set to 16k for GSM8k, Math-500, GPQA\_diamond, and 32k for AIME24. For AIME24 and GPQA\_diamond, due to their higher difficulty and smaller sizes, we sample 8 times and report the mean values. As for metrics, we adopt three primary metrics to assess both reasoning accuracy and compression effectiveness comprehensively.
\texttt{Acc.} denotes the accuracy of the final answer.
\texttt{Tok.} refers to the average response length, measured in tokens.
\texttt{CR} (Compression Rate) is defined as the ratio of the average response length to that of the original model, with lower values indicating better compression.

\paragraph{Baselines.}  
We compare our method against existing approaches that aim to remove redundant reasoning and construct efficient reasoning datasets, with further comparisons after fine-tuning under SFT and SimPO scenarios.
\textbf{OverThink} \cite{chen2024not} samples eight responses from the LRM with a relatively high temperature and selects the shortest correct one. Only the first reflection step after the answer is retained.
\textbf{Spirit} \cite{cui2025stepwise}  iteratively removes steps with minimal impact on perplexity (PPL), deleting 30\% of the least critical steps. Adjacent contexts are merged to maintain coherence. Appendix \ref{cha:training_details} shows more implementation and training details.

\subsection{Reults and Analysis}
This section evaluates different compression methods across various reasoning benchmarks. Our results show that \textsc{ConCISE} strikes a superior balance between compression and task performance compared to baseline methods, enabling models to effectively eliminate redundant reasoning steps. Additionally, models fine-tuned with \textsc{ConCISE} on mathematical datasets generalize well to GPQA\_diamond, showcasing its robustness.

\paragraph{\textsc{ConCISE} achieves superior balance between compression and task performance}
As shown in Table~\ref{tab:all_eval}, \textsc{ConCISE} consistently demonstrates strong compression ability across all four LRMs and under both SFT and SimPO settings, while maintaining model performance. Particularly under SimPO settings, \textsc{ConCISE} achieves a compression rate of \textasciitilde50\% with minimal impact on the original model's performance, achieving excellent compression results on the challenging task AIME24 as well as the out-of-domain task GPQA.

In contrast, OverThink achieves competitive task performance but exhibits weaker compression under both SFT and SimPO settings. Regarding Spirit, despite attaining a compression rate comparable to that of \textsc{ConCISE} under SimPO settings, its task performance suffered a notable degradation, especially on challenging tasks like AIME24. Overall, compared to baseline methods, \textsc{ConCISE} offers a better balance between compression and performance, demonstrating excellent in-domain and out-of-domain generalization and robustness.

\paragraph{How does \textsc{ConCISE} achieve a better compression \& performance balance?}
To better understand how \textsc{ConCISE} achieves an improved balance between compression and performance, we particularly analyze the structure of reasoning chains generated on various benchmarks of DeepSeek-R1-Distill-Qwen-7B. As shown in Figure~\ref{fig:steplen}, we evaluate the average number of steps (\texttt{StepNum}) and the average token count per step (\texttt{StepLen}). \textsc{ConCISE} generates the fewest reasoning steps, while its \texttt{StepLen} remains comparable to that of the original model.

More detailed analysis focuses on reflection steps within reasoning chains on MATH-500, revealing distinct compression strategies among the methods. As shown in Figure~\ref{fig:math_metric}, we find that \textbf{pruning reflection steps is key to compression, and preserving non-reflection steps is paramount for model performance.} \textsc{ConCISE} uniquely achieves this critical balance. Specifically, Spirit-SFT retains the most reflection steps (\texttt{RefNum}), followed by OverThink, while \textsc{ConCISE} retains the fewest. Regarding non-reflection steps (\texttt{Non-RefNum}), OverThink and \textsc{ConCISE} exhibit similar counts. Spirit shows slightly fewer non-reflection steps under SimPO, albeit at the cost of a performance drop. This suggests that current compression techniques primarily achieve compression by pruning reflection steps, as none of the evaluated methods can effectively reduce non-reflection steps without impairing model performance.

Further dissecting the reflection steps, OverThink and Spirit have the most reflection steps before FAS(\texttt{Pre-FAS RefNum}) due to ineffective early-stage reasoning control. In contrast, Spirit-SFT has most reflection steps after FAS(\texttt{Post-FAS RefNum}), indicating its inability to prune redundant post-answer reflections. These distinct strategies in reflection management contribute to performance disparities: OverThink's limited pre-answer pruning curtails its compression potential, while Spirit's difficulty in accurately identifying redundant steps often degrades performance by removing essential content. \textsc{ConCISE}, however, achieves a better compression-performance balance by precisely eliminating redundant reflection steps throughout the entire reasoning chain without compromising the essential reasoning process. We present more details about the above analysis in Appendix~\ref{cha:experiments_datails}

\paragraph{How does the training dataset affect compression?}
Since all methods operate under identical training configurations, variations in final compression performance are fundamentally rooted in the \textbf{characteristics of their respective training datasets}. To this end, we analyze training datasets constructed by \textsc{ConCISE}, OverThink, Spirit, Greedy-Search (derived from Spirit's rejected samples), and Longest-of-N (derived from rejected samples of \textsc{ConCISE} and OverThink). 

As shown in Figure~\ref{fig:train_metric}, surface-level metrics such as response tokens (\texttt{Tok.}), step count (\texttt{StepNum}), and average step length (\texttt{StepLen}) show no strong direct correlation with the final compression rate. For instance, while the training datasets of \textsc{ConCISE} and OverThink yield nearly identical values for these metrics, \textsc{ConCISE} demonstrates significantly better compression after fine-tuning under both SFT
and SimPO. Conversely, Metrics related to reflection offer more insightful distinctions among the methods. For instance, OverThink has the highest values for \texttt{Pre-FAS RefNum}, while Spirit shows the highest \texttt{Post-FAS RefNum}. \textsc{ConCISE}, in contrast, consistently maintains lower figures for both these metrics, underscoring its more effective control over the model's reflective tendencies. Echoing the analysis from the previous subsection, these observed characteristics within the training datasets directly mirror the fine-tuned model's capacity for reflection management. This provides strong evidence that the model effectively learns features about reflection from the training data during the fine-tuning process. Thus, it becomes clear how \textsc{ConCISE} uses its unique training data design to achieve significant compression by identifying and cutting down on unnecessary reflection steps.

\subsection{Ablation Study}

\begin{table}[t]
\centering
\resizebox{\linewidth}{!}{%
\begin{tabular}{p{1.6cm}l|cccc|c}
\toprule
\textbf{Benchmark} & \textbf{Method} & \textbf{Acc.} & \textbf{Tok.} & \textbf{StepNum} & \textbf{StepLen} & \textbf{CR} \\
\midrule
\multirow{3}{*}{Math-500}
& Origin        & 90.8 & 3854  & 113.7 & 33.9& 100\% \\
& \textbf{ConCISE}       & \underline{92.0} & \textbf{2244.3}  & 63.0  & 35.6 & \textbf{58\%} \\
& Conf. Inj.    & 91.8 & \underline{2620}  & 79.5  & 33.0 & \underline{68\%} \\
& Early Stop    & \textbf{92.4} & 2654  & 68.4  & 38.8 & 69\% \\
\midrule
\multirow{4}{*}{GSM8K}
& Origin        & 93.1 & 1442  & 41.4  & 34.8 & 100\% \\
& \textbf{ConCISE}       & \textbf{92.9} & \textbf{832}   & 22.5  & 37.1 & \textbf{58\%} \\
& Conf. Inj.    & 92.6 & \underline{934}   & 23.2  & 40.2 & \underline{65\%} \\
& Early Stop    & \underline{92.7} & 1003  & 22.3  & 45.0 & 70\% \\
\midrule
\multirow{4}{*}{AIME24}
& Origin        & 54.2 & 13574 & 463.0 & 29.3 & 100\% \\
& \textbf{ConCISE}       & \underline{52.1} & \textbf{9751} & 333.0 & 29.3 & \textbf{72\%} \\
& Conf. Inj.    & 51.3 & \underline{10166} & 372.2 & 27.3 & \underline{75\%} \\
& Early Stop    & \textbf{54.2} & 12205 & 389.7 & 31.3 & 90\% \\
\bottomrule
\end{tabular}%
}
\caption{Ablation study of Confidence Injection and Early Stopping on DeepSeek-R1-Distill-Qwen-7B.}
\label{tab:ablation-combined}
\end{table}

We conduct an ablation study to evaluate the two core components of \textsc{ConCISE}: \textit{Confidence Injection} and \textit{Early Stopping}. For each, we construct a training dataset and fine-tune DeepSeek-7B using the same configurations as in the main experiments. All experiments are performed under SFT settings and evaluated across three mathematical reasoning tasks: Math-500, GSM8K, and AIME24. 

The results, summarized in Table~\ref{tab:ablation-combined}, show that both mechanisms achieve similar reasoning accuracy to full \textsc{ConCISE} but exhibit weaker compression performance. Specifically, \textit{Confidence Injection} suppresses unnecessary reflections by boosting the model's confidence, but has a limited impact on terminating reasoning after the final answer. \textit{Early Stopping}, on the other hand, halts excessive post-answer reasoning but does not address earlier stages. Only by combining both can we achieve comprehensive redundancy elimination without compromising reasoning quality.

\begin{figure}[t]
\centering
\includegraphics[width=\linewidth]{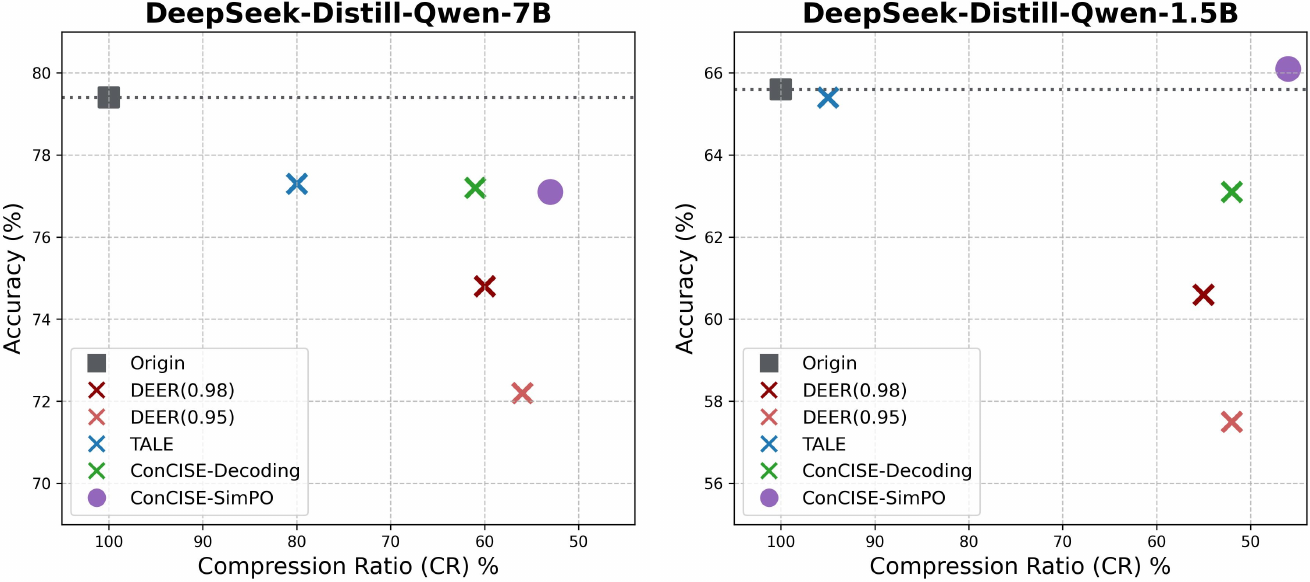}
\caption{Average performance of training-free methods and \textsc{ConCISE} variants on DeepSeek-R1-Distill-Qwen-7B and 1.5B across GSM8K, Math-500, and AIME24.}
\label{fig:training_free}
\end{figure}

\subsection{Analysis of Training-free Methods}
\label{sec:trainingfree}

To further demonstrate the advantages of the \textsc{ConCISE} framework, we compare it against representative \textit{training-free} methods. 
Specifically, we include \textbf{TALE}~\cite{han2024token}, which constrains reasoning length by injecting explicit token budgets into prompts, 
and \textbf{DEER}~\cite{yang2025dynamic}, which dynamically decides when to terminate reasoning by monitoring the generation probability of the answer. 
For DEER, we reproduce two variants with $\lambda=0.95$ and $\lambda=0.98$. 
We also implement a training-free version of our method, denoted \textbf{ConCISE-Decoding}, 
which directly applies Algorithm~\ref{alg: concise}during inference to decide when to terminate reasoning, without relying on ground-truth verification. 
We evaluate these methods on GSM8K, Math-500, and AIME24, and report the averaged results in Figure~\ref{fig:training_free}, 
with detailed statistics provided in Table~\ref{tab:trainingfree}.

We find that training-free methods can achieve partial compression, but their compression levels are unstable 
and often come with substantial accuracy degradation. 
For instance, DEER achieves relatively stable compression but introduces a 5--8\% accuracy drop. 
Meanwhile, TALE attains a high compression ratio on GSM8K with DeepSeek-7B, but performs poorly on other datasets and models. 
In contrast, \textsc{ConCISE-SimPO} consistently achieves a superior overall trade-off, and even the training-free variant \textsc{ConCISE-Decoding} outperforms other training-free approaches. 
These results highlight that \textsc{ConCISE} enables more precise elimination of redundant reflection steps, thereby yielding higher-quality reasoning chains.
\section{Conclusion}

In this work, we first modeled the generation of reflection steps in LRMs from a confidence-guided perspective, helping to understand two key patterns responsible for redundant reflection within the reasoning process of LRMs: \textit{Confidence Deficit} and \textit{Termination Delay}. Based on this, we propose \textsc{ConCISE}, which employs \textit{Confidence Injection} and \textit{Early Stopping} mechanisms to respectively address the above two patterns to suppress redundant reflection steps and produce efficient, concise reasoning chains. Extensive experiments demonstrate that LRMs fine-tuned on \textsc{ConCISE}-generated data significantly compress their responses while maintaining strong task performance.

\section*{Limitations}

While \textsc{ConCISE} demonstrates strong performance in compressing reasoning chains, it is subject to certain limitations that merit further exploration:

\paragraph{Room for Further Compression.}
While \textsc{ConCISE} effectively addresses redundant reflection steps, analysis indicates that significant potential remains for further compression of both non-reflection steps and the average length of each step. Future work could explore integrating techniques with \textsc{ConCISE} to achieve a higher degree of compression while maintaining model performance.

\paragraph{Confidence Estimation During Reasoning.}
Although \textsc{ConCISE} introduces a lightweight confidence detector for the post-answer phase, it still lacks a direct mechanism to model confidence before the first answer. Instead, it relies on the occurrence of reflection steps as an indirect proxy. Future work could explore training a lightweight model to detect the model's internal confidence, thereby enabling more fine-grained control.

\paragraph{Integration with RLVR methods.}
This work focuses on constructing concise reasoning traces and validating them under SFT and SimPO, which we consider sufficient to support our core contribution. 
At the same time, we recognize that RLVR methods such as GRPO\cite{shao2024deepseekmath} are powerful and widely used in post-training. Exploring how confidence-guided signals can be incorporated into RLVR setups is a valuable future direction.
\section*{Acknowledgement}
This research was supported in part by the National Natural Science Foundation of China under Grant No. 62402267, 62432004, the China Postdoctoral Science Foundation under Grant No. BX20240177, 2025M771499, and a grant from the Guoqiang Institute, Tsinghua University.
\bibliography{main}

\begin{thebibliography}{45}
\providecommand{\natexlab}[1]{#1}

\bibitem[{Aggarwal and Welleck(2025)}]{aggarwal2025l1}
Pranjal Aggarwal and Sean Welleck. 2025.
\newblock L1: Controlling how long a reasoning model thinks with reinforcement learning.
\newblock \emph{arXiv preprint arXiv:2503.04697}.

\bibitem[{Arora and Zanette(2025)}]{arora2025training}
Daman Arora and Andrea Zanette. 2025.
\newblock Training language models to reason efficiently.
\newblock \emph{arXiv preprint arXiv:2502.04463}.

\bibitem[{Aytes et~al.(2025)Aytes, Baek, and Hwang}]{aytes2025sketch}
Simon~A Aytes, Jinheon Baek, and Sung~Ju Hwang. 2025.
\newblock Sketch-of-thought: Efficient llm reasoning with adaptive cognitive-inspired sketching.
\newblock \emph{arXiv preprint arXiv:2503.05179}.

\bibitem[{Chen et~al.(2024)Chen, Xu, Liang, He, Pang, Yu, Song, Liu, Zhou, Zhang et~al.}]{chen2024not}
Xingyu Chen, Jiahao Xu, Tian Liang, Zhiwei He, Jianhui Pang, Dian Yu, Linfeng Song, Qiuzhi Liu, Mengfei Zhou, Zhuosheng Zhang, and 1 others. 2024.
\newblock Do not think that much for 2+ 3=? on the overthinking of o1-like llms.
\newblock \emph{arXiv preprint arXiv:2412.21187}.

\bibitem[{Chuang et~al.(2025)Chuang, Yu, Wang, Zhang, Liu, Cai, Sui, Braverman, and Hu}]{chuang2025confident}
Yu-Neng Chuang, Leisheng Yu, Guanchu Wang, Lizhe Zhang, Zirui Liu, Xuanting Cai, Yang Sui, Vladimir Braverman, and Xia Hu. 2025.
\newblock Confident or seek stronger: Exploring uncertainty-based on-device llm routing from benchmarking to generalization.
\newblock \emph{arXiv preprint arXiv:2502.04428}.

\bibitem[{Cobbe et~al.(2021)Cobbe, Kosaraju, Bavarian, Chen, Jun, Kaiser, Plappert, Tworek, Hilton, Nakano et~al.}]{cobbe2021training}
Karl Cobbe, Vineet Kosaraju, Mohammad Bavarian, Mark Chen, Heewoo Jun, Lukasz Kaiser, Matthias Plappert, Jerry Tworek, Jacob Hilton, Reiichiro Nakano, and 1 others. 2021.
\newblock Training verifiers to solve math word problems.
\newblock \emph{arXiv preprint arXiv:2110.14168}.

\bibitem[{Cui et~al.(2025)Cui, He, Zeng, Liu, Tang, Dai, Han, Luo, Huang, Li et~al.}]{cui2025stepwise}
Yingqian Cui, Pengfei He, Jingying Zeng, Hui Liu, Xianfeng Tang, Zhenwei Dai, Yan Han, Chen Luo, Jing Huang, Zhen Li, and 1 others. 2025.
\newblock Stepwise perplexity-guided refinement for efficient chain-of-thought reasoning in large language models.
\newblock \emph{arXiv preprint arXiv:2502.13260}.

\bibitem[{Feng et~al.(2025)Feng, Fang, Ma, and Wang}]{feng2025efficient}
Sicheng Feng, Gongfan Fang, Xinyin Ma, and Xinchao Wang. 2025.
\newblock Efficient reasoning models: A survey.
\newblock \emph{arXiv preprint arXiv:2504.10903}.

\bibitem[{Guo et~al.(2025)Guo, Yang, Zhang, Song, Zhang, Xu, Zhu, Ma, Wang, Bi et~al.}]{guo2025deepseek}
Daya Guo, Dejian Yang, Haowei Zhang, Junxiao Song, Ruoyu Zhang, Runxin Xu, Qihao Zhu, Shirong Ma, Peiyi Wang, Xiao Bi, and 1 others. 2025.
\newblock Deepseek-r1: Incentivizing reasoning capability in llms via reinforcement learning.
\newblock \emph{arXiv preprint arXiv:2501.12948}.

\bibitem[{Han et~al.(2024)Han, Wang, Fang, Zhao, Ma, and Chen}]{han2024token}
Tingxu Han, Zhenting Wang, Chunrong Fang, Shiyu Zhao, Shiqing Ma, and Zhenyu Chen. 2024.
\newblock Token-budget-aware llm reasoning.
\newblock \emph{arXiv preprint arXiv:2412.18547}.

\bibitem[{Hao et~al.(2024)Hao, Sukhbaatar, Su, Li, Hu, Weston, and Tian}]{hao2024training}
Shibo Hao, Sainbayar Sukhbaatar, DiJia Su, Xian Li, Zhiting Hu, Jason Weston, and Yuandong Tian. 2024.
\newblock Training large language models to reason in a continuous latent space.
\newblock \emph{arXiv preprint arXiv:2412.06769}.

\bibitem[{He et~al.(2025)He, Liu, Liu, Yan, Wang, Cheng, Zhang, Zhang, Xu, Shen, Li, Zeng, Wei, Cheng, An, Liu, and Zhou}]{skywork-or1-2025}
Jujie He, Jiacai Liu, Chris~Yuhao Liu, Rui Yan, Chaojie Wang, Peng Cheng, Xiaoyu Zhang, Fuxiang Zhang, Jiacheng Xu, Wei Shen, Siyuan Li, Liang Zeng, Tianwen Wei, Cheng Cheng, Bo~An, Yang Liu, and Yahui Zhou. 2025.
\newblock Skywork open reasoner series.
\newblock \url{https://capricious-hydrogen-41c.notion.site/Skywork-Open-Reaonser-Series-1d0bc9ae823a80459b46c149e4f51680}.
\newblock Notion Blog.

\bibitem[{Hendrycks et~al.(2021)Hendrycks, Burns, Kadavath, Arora, Basart, Tang, Song, and Steinhardt}]{hendrycksmath2021}
Dan Hendrycks, Collin Burns, Saurav Kadavath, Akul Arora, Steven Basart, Eric Tang, Dawn Song, and Jacob Steinhardt. 2021.
\newblock Measuring mathematical problem solving with the math dataset.
\newblock \emph{NeurIPS}.

\bibitem[{Jaech et~al.(2024)Jaech, Kalai, Lerer, Richardson, El-Kishky, Low, Helyar, Madry, Beutel, Carney et~al.}]{jaech2024openai}
Aaron Jaech, Adam Kalai, Adam Lerer, Adam Richardson, Ahmed El-Kishky, Aiden Low, Alec Helyar, Aleksander Madry, Alex Beutel, Alex Carney, and 1 others. 2024.
\newblock Openai o1 system card.
\newblock \emph{arXiv preprint arXiv:2412.16720}.

\bibitem[{Lee et~al.(2025)Lee, Che, and Peng}]{lee2025well}
Ayeong Lee, Ethan Che, and Tianyi Peng. 2025.
\newblock How well do llms compress their own chain-of-thought? a token complexity approach.
\newblock \emph{arXiv preprint arXiv:2503.01141}.

\bibitem[{Luo et~al.(2025)Luo, Shen, He, Wang, Liu, Li, Tan, Cao, and Tao}]{luo2025o1}
Haotian Luo, Li~Shen, Haiying He, Yibo Wang, Shiwei Liu, Wei Li, Naiqiang Tan, Xiaochun Cao, and Dacheng Tao. 2025.
\newblock O1-pruner: Length-harmonizing fine-tuning for o1-like reasoning pruning.
\newblock \emph{arXiv preprint arXiv:2501.12570}.

\bibitem[{Ma et~al.(2025)Ma, Wan, Yu, Fang, and Wang}]{ma2025cot}
Xinyin Ma, Guangnian Wan, Runpeng Yu, Gongfan Fang, and Xinchao Wang. 2025.
\newblock Cot-valve: Length-compressible chain-of-thought tuning.
\newblock \emph{arXiv preprint arXiv:2502.09601}.

\bibitem[{Meng et~al.(2024)Meng, Xia, and Chen}]{meng2024simpo}
Yu~Meng, Mengzhou Xia, and Danqi Chen. 2024.
\newblock Simpo: Simple preference optimization with a reference-free reward.
\newblock \emph{Advances in Neural Information Processing Systems}, 37:124198--124235.

\bibitem[{Muennighoff et~al.(2025)Muennighoff, Yang, Shi, Li, Fei-Fei, Hajishirzi, Zettlemoyer, Liang, Cand{\`e}s, and Hashimoto}]{muennighoff2025s1}
Niklas Muennighoff, Zitong Yang, Weijia Shi, Xiang~Lisa Li, Li~Fei-Fei, Hannaneh Hajishirzi, Luke Zettlemoyer, Percy Liang, Emmanuel Cand{\`e}s, and Tatsunori Hashimoto. 2025.
\newblock s1: Simple test-time scaling.
\newblock \emph{arXiv preprint arXiv:2501.19393}.

\bibitem[{Munkhbat et~al.(2025)Munkhbat, Ho, Kim, Yang, Kim, and Yun}]{munkhbat2025self}
Tergel Munkhbat, Namgyu Ho, Seo~Hyun Kim, Yongjin Yang, Yujin Kim, and Se-Young Yun. 2025.
\newblock Self-training elicits concise reasoning in large language models.
\newblock \emph{arXiv preprint arXiv:2502.20122}.

\bibitem[{Nayab et~al.(2024)Nayab, Rossolini, Simoni, Saracino, Buttazzo, Manes, and Giacomelli}]{nayab2024concise}
Sania Nayab, Giulio Rossolini, Marco Simoni, Andrea Saracino, Giorgio Buttazzo, Nicolamaria Manes, and Fabrizio Giacomelli. 2024.
\newblock Concise thoughts: Impact of output length on llm reasoning and cost.
\newblock \emph{arXiv preprint arXiv:2407.19825}.

\bibitem[{Ong et~al.(2024)Ong, Almahairi, Wu, Chiang, Wu, Gonzalez, Kadous, and Stoica}]{ong2024routellm}
Isaac Ong, Amjad Almahairi, Vincent Wu, Wei-Lin Chiang, Tianhao Wu, Joseph~E Gonzalez, M~Waleed Kadous, and Ion Stoica. 2024.
\newblock Routellm: Learning to route llms from preference data.
\newblock In \emph{The Thirteenth International Conference on Learning Representations}.

\bibitem[{Qu et~al.(2025)Qu, Li, Su, Sun, Yan, Liu, Cui, Liu, Liang, He et~al.}]{qu2025survey}
Xiaoye Qu, Yafu Li, Zhaochen Su, Weigao Sun, Jianhao Yan, Dongrui Liu, Ganqu Cui, Daizong Liu, Shuxian Liang, Junxian He, and 1 others. 2025.
\newblock A survey of efficient reasoning for large reasoning models: Language, multimodality, and beyond.
\newblock \emph{arXiv preprint arXiv:2503.21614}.

\bibitem[{{Qwen Team}(2024)}]{qwq-32b-preview}
{Qwen Team}. 2024.
\newblock \href{https://qwenlm.github.io/blog/qwq-32b-preview/}{QwQ: Reflect Deeply on the Boundaries of the Unknown}.
\newblock Accessed: 2025-04-05.

\bibitem[{Rafailov et~al.(2023)Rafailov, Sharma, Mitchell, Manning, Ermon, and Finn}]{rafailov2023direct}
Rafael Rafailov, Archit Sharma, Eric Mitchell, Christopher~D Manning, Stefano Ermon, and Chelsea Finn. 2023.
\newblock Direct preference optimization: Your language model is secretly a reward model.
\newblock \emph{Advances in Neural Information Processing Systems}, 36:53728--53741.

\bibitem[{Rein et~al.(2024)Rein, Hou, Stickland, Petty, Pang, Dirani, Michael, and Bowman}]{rein2024gpqa}
David Rein, Betty~Li Hou, Asa~Cooper Stickland, Jackson Petty, Richard~Yuanzhe Pang, Julien Dirani, Julian Michael, and Samuel~R Bowman. 2024.
\newblock Gpqa: A graduate-level google-proof q\&a benchmark.
\newblock In \emph{First Conference on Language Modeling}.

\bibitem[{Renze and Guven(2024)}]{renze2024benefits}
Matthew Renze and Erhan Guven. 2024.
\newblock The benefits of a concise chain of thought on problem-solving in large language models.
\newblock In \emph{2024 2nd International Conference on Foundation and Large Language Models (FLLM)}, pages 476--483. IEEE.

\bibitem[{Shao et~al.(2024)Shao, Wang, Zhu, Xu, Song, Bi, Zhang, Zhang, Li, Wu et~al.}]{shao2024deepseekmath}
Zhihong Shao, Peiyi Wang, Qihao Zhu, Runxin Xu, Junxiao Song, Xiao Bi, Haowei Zhang, Mingchuan Zhang, YK~Li, Yang Wu, and 1 others. 2024.
\newblock Deepseekmath: Pushing the limits of mathematical reasoning in open language models.
\newblock \emph{arXiv preprint arXiv:2402.03300}.

\bibitem[{Shen et~al.(2025{\natexlab{a}})Shen, Zhang, Huang, Shi, Zhang, Yan, Wang, Wang, and Lian}]{shen2025dast}
Yi~Shen, Jian Zhang, Jieyun Huang, Shuming Shi, Wenjing Zhang, Jiangze Yan, Ning Wang, Kai Wang, and Shiguo Lian. 2025{\natexlab{a}}.
\newblock Dast: Difficulty-adaptive slow-thinking for large reasoning models.
\newblock \emph{arXiv preprint arXiv:2503.04472}.

\bibitem[{Shen et~al.(2025{\natexlab{b}})Shen, Yan, Zhang, Hu, Du, and He}]{shen2025codi}
Zhenyi Shen, Hanqi Yan, Linhai Zhang, Zhanghao Hu, Yali Du, and Yulan He. 2025{\natexlab{b}}.
\newblock Codi: Compressing chain-of-thought into continuous space via self-distillation.
\newblock \emph{arXiv preprint arXiv:2502.21074}.

\bibitem[{Sui et~al.(2025)Sui, Chuang, Wang, Zhang, Zhang, Yuan, Liu, Wen, Chen, Hu et~al.}]{sui2025stop}
Yang Sui, Yu-Neng Chuang, Guanchu Wang, Jiamu Zhang, Tianyi Zhang, Jiayi Yuan, Hongyi Liu, Andrew Wen, Hanjie Chen, Xia Hu, and 1 others. 2025.
\newblock Stop overthinking: A survey on efficient reasoning for large language models.
\newblock \emph{arXiv preprint arXiv:2503.16419}.

\bibitem[{Sun et~al.(2024)Sun, Haider, Zhang, Yang, Qiu, Yin, Wang, Bartlett, and Zanette}]{sun2024fast}
Hanshi Sun, Momin Haider, Ruiqi Zhang, Huitao Yang, Jiahao Qiu, Ming Yin, Mengdi Wang, Peter Bartlett, and Andrea Zanette. 2024.
\newblock Fast best-of-n decoding via speculative rejection.
\newblock \emph{arXiv preprint arXiv:2410.20290}.

\bibitem[{Team et~al.(2025)Team, Du, Gao, Xing, Jiang, Chen, Li, Xiao, Du, Liao et~al.}]{team2025kimi}
Kimi Team, Angang Du, Bofei Gao, Bowei Xing, Changjiu Jiang, Cheng Chen, Cheng Li, Chenjun Xiao, Chenzhuang Du, Chonghua Liao, and 1 others. 2025.
\newblock Kimi k1. 5: Scaling reinforcement learning with llms.
\newblock \emph{arXiv preprint arXiv:2501.12599}.

\bibitem[{Team(2025)}]{qwen3}
Qwen Team. 2025.
\newblock \href {https://qwenlm.github.io/blog/qwen3/} {Qwen3}.

\bibitem[{Wang et~al.(2025)Wang, Liu, Xu, Liang, Chen, He, Song, Yu, Li, Zhang et~al.}]{wang2025thoughts}
Yue Wang, Qiuzhi Liu, Jiahao Xu, Tian Liang, Xingyu Chen, Zhiwei He, Linfeng Song, Dian Yu, Juntao Li, Zhuosheng Zhang, and 1 others. 2025.
\newblock Thoughts are all over the place: On the underthinking of o1-like llms.
\newblock \emph{arXiv preprint arXiv:2501.18585}.

\bibitem[{Wei et~al.(2021)Wei, Bosma, Zhao, Guu, Yu, Lester, Du, Dai, and Le}]{wei2021finetuned}
Jason Wei, Maarten Bosma, Vincent~Y Zhao, Kelvin Guu, Adams~Wei Yu, Brian Lester, Nan Du, Andrew~M Dai, and Quoc~V Le. 2021.
\newblock Finetuned language models are zero-shot learners.
\newblock \emph{arXiv preprint arXiv:2109.01652}.

\bibitem[{Wei et~al.(2022)Wei, Wang, Schuurmans, Bosma, Xia, Chi, Le, Zhou et~al.}]{wei2022chain}
Jason Wei, Xuezhi Wang, Dale Schuurmans, Maarten Bosma, Fei Xia, Ed~Chi, Quoc~V Le, Denny Zhou, and 1 others. 2022.
\newblock Chain-of-thought prompting elicits reasoning in large language models.
\newblock \emph{Advances in neural information processing systems}, 35:24824--24837.

\bibitem[{Wu et~al.(2025)Wu, Wang, Du, Jegelka, and Wang}]{wu2025more}
Yuyang Wu, Yifei Wang, Tianqi Du, Stefanie Jegelka, and Yisen Wang. 2025.
\newblock When more is less: Understanding chain-of-thought length in llms.
\newblock \emph{arXiv preprint arXiv:2502.07266}.

\bibitem[{Xia et~al.(2025)Xia, Li, Leong, Wang, and Li}]{xia2025tokenskip}
Heming Xia, Yongqi Li, Chak~Tou Leong, Wenjie Wang, and Wenjie Li. 2025.
\newblock Tokenskip: Controllable chain-of-thought compression in llms.
\newblock \emph{arXiv preprint arXiv:2502.12067}.

\bibitem[{Xu et~al.(2025)Xu, Xie, Zhao, and He}]{xu2025chain}
Silei Xu, Wenhao Xie, Lingxiao Zhao, and Pengcheng He. 2025.
\newblock Chain of draft: Thinking faster by writing less.
\newblock \emph{arXiv preprint arXiv:2502.18600}.

\bibitem[{Yang et~al.(2024)Yang, Yang, Zhang, Hui, Zheng, Yu, Li, Liu, Huang, Wei, Lin, Yang, Tu, Zhang, Yang, Yang, Zhou, Lin, Dang, Lu, Bao, Yang, Yu, Li, Xue, Zhang, Zhu, Men, Lin, Li, Xia, Ren, Ren, Fan, Su, Zhang, Wan, Liu, Cui, Zhang, and Qiu}]{qwen2.5}
An~Yang, Baosong Yang, Beichen Zhang, Binyuan Hui, Bo~Zheng, Bowen Yu, Chengyuan Li, Dayiheng Liu, Fei Huang, Haoran Wei, Huan Lin, Jian Yang, Jianhong Tu, Jianwei Zhang, Jianxin Yang, Jiaxi Yang, Jingren Zhou, Junyang Lin, Kai Dang, and 22 others. 2024.
\newblock Qwen2.5 technical report.
\newblock \emph{arXiv preprint arXiv:2412.15115}.

\bibitem[{Yang et~al.(2025)Yang, Si, Duan, Zhu, Zhu, Lin, Cao, and Wang}]{yang2025dynamic}
Chenxu Yang, Qingyi Si, Yongjie Duan, Zheliang Zhu, Chenyu Zhu, Zheng Lin, Li~Cao, and Weiping Wang. 2025.
\newblock Dynamic early exit in reasoning models.
\newblock \emph{arXiv preprint arXiv:2504.15895}.

\bibitem[{Yu et~al.(2025)Yu, Zhang, Zhu, Yuan, Zuo, Yue, Fan, Liu, Liu, Liu et~al.}]{yu2025dapo}
Qiying Yu, Zheng Zhang, Ruofei Zhu, Yufeng Yuan, Xiaochen Zuo, Yu~Yue, Tiantian Fan, Gaohong Liu, Lingjun Liu, Xin Liu, and 1 others. 2025.
\newblock Dapo: An open-source llm reinforcement learning system at scale.
\newblock \emph{arXiv preprint arXiv:2503.14476}.

\bibitem[{Zhang et~al.(2025)Zhang, Zhu, Sun, Luo, Qiao, Du, Zheng, Chen, and Zhang}]{zhang2025lightthinker}
Jintian Zhang, Yuqi Zhu, Mengshu Sun, Yujie Luo, Shuofei Qiao, Lun Du, Da~Zheng, Huajun Chen, and Ningyu Zhang. 2025.
\newblock Lightthinker: Thinking step-by-step compression.
\newblock \emph{arXiv preprint arXiv:2502.15589}.

\bibitem[{Zheng et~al.(2024)Zheng, Zhang, Zhang, Ye, Luo, Feng, and Ma}]{zheng2024llamafactory}
Yaowei Zheng, Richong Zhang, Junhao Zhang, Yanhan Ye, Zheyan Luo, Zhangchi Feng, and Yongqiang Ma. 2024.
\newblock \href {http://arxiv.org/abs/2403.13372} {Llamafactory: Unified efficient fine-tuning of 100+ language models}.
\newblock In \emph{Proceedings of the 62nd Annual Meeting of the Association for Computational Linguistics (Volume 3: System Demonstrations)}, Bangkok, Thailand. Association for Computational Linguistics.

\end{thebibliography}
\cleardoublepage
\appendix

\section{Details about ConCISE}
\label{sec:appendix_detail}

\subsection{Confidence Injection}
\label{cha:conf_pool}
The selection of appropriate confidence phrases is critical for the Confidence Injection component of ConCISE. We curated a pool of 20 distinct phrases, presented in Table~\ref{tab:confphrase}, by observing the model's native reasoning expressions, employing manual design, and performing experimental refinement (partly illustrated in Figure~\ref{fig:refprob}). The purpose of these phrases is to inject confidence into the model, thereby preventing redundant reflection steps during reasoning. To mitigate potential overfitting to any specific phrase, one phrase is randomly selected from the pool for injection at runtime.

Additionally, another crucial mechanism within the Confidence Injection process is identifying whether the current step constitutes a reflection step. The accurate and swift identification of such steps is paramount for the success of Confidence Injection. Consequently, informed by our observations of reflection steps during LRM reasoning processes, we employed a rule-based detection method. This method, which is detailed in Appendix~\ref{eq:rule_based_detection}, achieves reasonably accurate detection while having a negligible impact on the model's inference speed.

\begin{table}[htbp]
\centering
\caption{The Pool of 20 Confidence Phrases used for Confidence Injection.}
\label{tab:confphrase}
\begin{tabular}{cl}
\toprule
Index & Confidence Phrase \\
\midrule
1 & Therefore \\
2 & The reasoning holds \\
3 & Previous steps are correct \\
4 & All steps are valid \\
5 & With this established \\
6 & That sounds reasonable \\
7 & Let's go ahead \\
8 & Alright, let's carry on \\
9 & Let's proceed \\
10 & Let’s progress \\
11 & So, putting it all together \\ 
12 & The logic stands firm \\
13 & The reasoning process is valid \\
14 & Good, let's keep going\\
15 & Everything seems reasonable so far \\
16 & This part checks out \\
17 & I think that's solid. So \\
18 & The reasoning holds, let's keep going\\
19 & Everything checks out, let's move on \\
20 & All steps are solid, let's move forward \\ 
\bottomrule
\end{tabular}
\end{table}
\subsection{Early Stopping}
\label{cha:early_prompt}
\paragraph{Probing prompt and confidence-indicated tokens.}
The effectiveness of the Early Stopping mechanism relies heavily on the accuracy of its confidence detector. To identify an appropriate probing prompt, we first analyzed the reasoning chains generated by four Large Reasoning Models (LRMs) using Greedy decoding on the training question set. This analysis revealed that the model frequently expresses confidence during the later part of the reasoning chain, and as shown in Table~\ref{tab:confidence_phrase_freq}, self-reported confidence expressions starting with ``I'm'' (such as ``I'm confident'') are very common.

Based on this observation, particularly considering the prevalence of confidence statements led by ``I'm'', we considered \texttt{"So, I'm"} as a potential probing prompt. Concurrently, we also experimented with alternative probing prompts, such as \texttt{"So, I feel"} and \texttt{"I can be"}. However, we observed that these alternative prompts often resulted in high confidence scores before the model had reached the final answer (i.e., prior to Final Answer Serialization, FAS), leading to unnecessary computational overhead.

Therefore, considering these factors—namely, the prevalence of ``I'm''-led confidence statements in the model's natural expressions and the shortcomings of the alternative prompts—we ultimately selected \texttt{"So, I'm"} as the fixed probing prompt. Subsequently, we identified the primary confidence-indicative tokens that follow the \texttt{"So, I'm"} prompt, which are: \texttt{"confident"}, \texttt{"pretty confident"}, \texttt{"sure"}, and \texttt{"pretty sure"}.

\begin{table*}[htbp]
\centering
\caption{Frequency of Common Confidence Expressions Observed in 2000 training Reasoning Chains Across Different Large Reasoning Model Series(DeepSeek, Qwen and Skywork)}
\label{tab:confidence_phrase_freq}
\resizebox{\textwidth}{!}{
\begin{tabular}{lrrrr}
\toprule
Expression & DeepSeek-7B & DeepSeek-1.5B & Qwen3-8B & Skywork-OR17B \\
\midrule
I'm confident & 423 & 335 & 127 & 560 \\
I feel confident & 170 & 172 & 58 & 144 \\
I can be confident that & 157 & 156 & 475 & 130 \\
I can confidently say & 70 & 176 & 91 & 98 \\
This gives me (more)confidence that& 68 & 81 & 23 & 86 \\
I'm pretty confident & 43 & 69 & 2 & 60 \\
I'm pretty sure & 37 & 28 & 2 & 17 \\
Therefore, confident that & 1 & 0 & 26 & 5\\
\midrule
\textbf{Total (sum of listed expressions)} & \textbf{969} & \textbf{1017} & \textbf{804} & \textbf{1100} \\
\bottomrule
\end{tabular}
}
\end{table*}

The confidence score, $\hat{c}_i$, for a given state $S_i$ is then calculated by summing the probabilities of generating these tokens or sequences immediately following the probing prompt \texttt{"So, I'm"}:
$$
\begin{aligned}
\hat{c}_i = & P(\text{"confident"} \mid S_i, \text{"So, I'm"})\\
& + P(\text{"sure"} \mid S_i, \text{"So, I'm"}) \\
& + P(\text{"pretty"} \mid S_i, \text{"So, I'm"}) \times \\
& \big( P(\text{"confident"} \mid S_i, \text{"So, I'm pretty"}) \\
& \quad + P(\text{"sure"} \mid S_i, \text{"So, I'm pretty"}) \big)
\end{aligned}
$$

Here, notation like $P(\text{token} \mid S_i, \text{sequence})$ represents the conditional probability of generating the next \texttt{token} given the preceding state $S_i$ and the specific text \texttt{sequence} provided as context. As demonstrated in Figure~\ref{fig:confidence} and supported by our experimental results, the confidence detector designed using this probing prompt and set of confidence-indicative tokens proves effective for the early stopping mechanism in ConCISE.

\paragraph{Early stopping threshold.}
The selection of the early stopping threshold, $t$, was guided by the principle of tailoring the reflection process to the apparent quality of the reasoning chain. Our objective was to configure the threshold such that:
\begin{itemize}
    \item Simple problems with rigorous reasoning steps require no post-answer reflection.
    \item Problems of moderate difficulty benefit from exactly one round of reflection.
    \item Difficult problems or those with less rigorous reasoning undergo two or more rounds of reflection.
\end{itemize}

\begin{figure}[htbp]
    \centering
    \includegraphics[width=\linewidth]{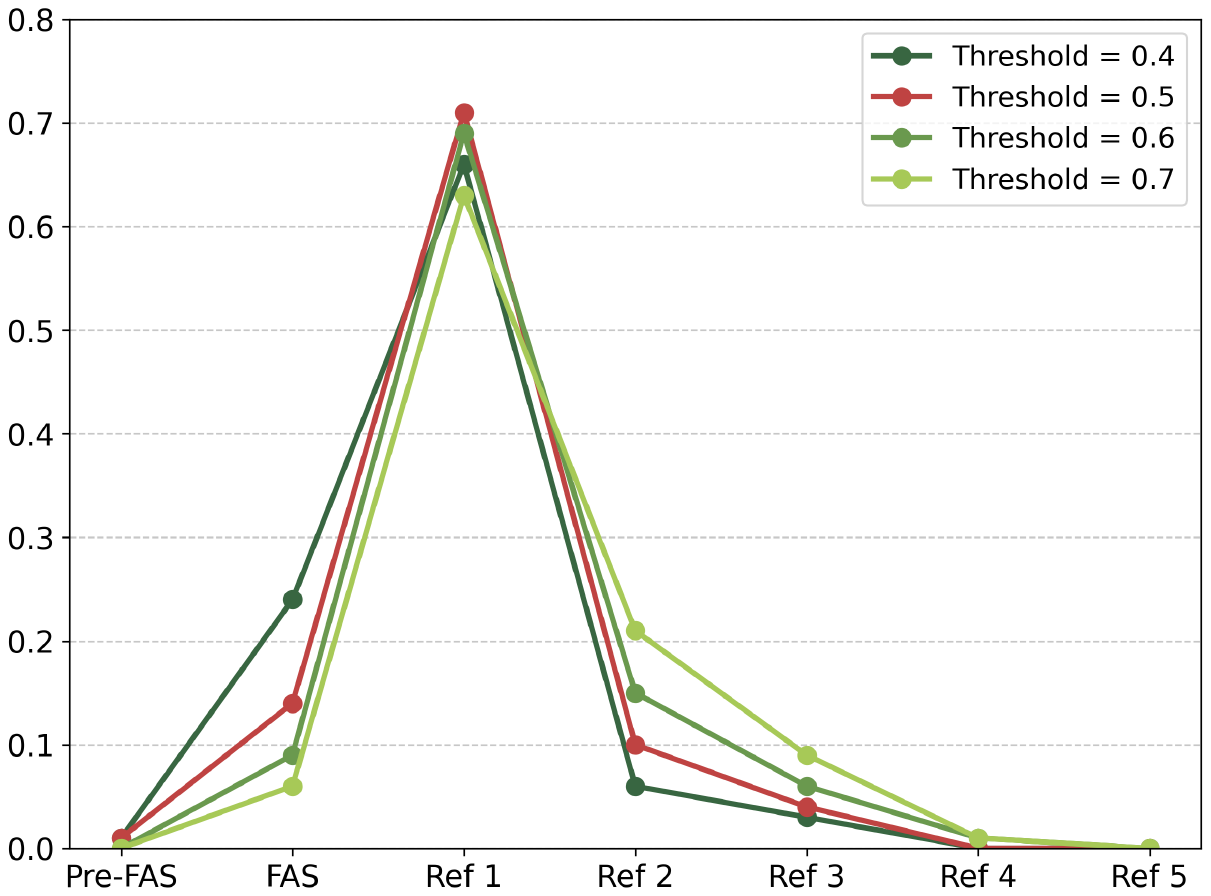}
    \caption{Distribution of exceeding points(when $\hat{c}_i>t_e$) probability of Pre-FAS, FAS, Ref1, 2, ..., 5 for different threshold values ($t_e$).}
    \label{fig:earlyratio}
\end{figure}

To find a threshold value that aligns with this principle, we empirically evaluated $t_e \in \{0.4, 0.5, 0.6, 0.7\}$ on the DeepSeek-R1-Distill-Qwen-7B model. For each value, we measured the probability distribution of the exceeding points(when $\hat{c}_i>t_e$) across different stages: exceeding after the initial answer generation (0 reflections), after the first reflection, after the second reflection, etc. The results of this analysis are illustrated in Figure~\ref{fig:earlyratio}.

\begin{table}[htbp]
\centering

\small
\renewcommand{\arraystretch}{1.15}
\resizebox{\linewidth}{!}{%
\begin{tabular}{p{1.6cm}l|cccc|c}
\toprule
\textbf{Benchmark} & \textbf{$T_e$} & \textbf{Acc.} & \textbf{Tok.} & \textbf{StepNum} & \textbf{StepLen} &\textbf{CR} \\
\midrule
\multirow{4}{*}{Math-500}
& 0.4 & 91.4 & 2390.6& 68.4 & 34.9 & 62.0\% \\
& \textbf{0.5} & \textbf{92.0} & \textbf{2244.3}  & \textbf{63.0} &\textbf{35.6} & \textbf{58.2\%} \\
& 0.6 & 91.8 & 2324.3 & 66.1 & 35.2 & 60.3\% \\
& 0.7 & 92.0 & 2345.1 & 67.5 & 34.8 &  60.8\% \\
\midrule
\multirow{4}{*}{GSM8K}
& 0.4  & 92.4 & 828.9 & 19.9 & 41.6 & 57.5\% \\
& \textbf{0.5}  & \textbf{92.9} & \textbf{831.9} & \textbf{22.5} &\textbf{37.1}  & \textbf{57.7\%} \\
& 0.6   & 93.0 & 831.7 & 20.4 & 40.7 & 57.7\% \\
& 0.7   & 92.7 & 849.3  & 21.7 & 39.2 & 58.9\% \\
\midrule
\multirow{4}{*}{AIME24}
& 0.4   & 45.8 & 11548.7 & 402.1 & 28.7 & 85.1\% \\
& \textbf{0.5} & \textbf{52.1} & \textbf{9750.8} & \textbf{333.0} &\textbf{29.3}  & \textbf{71.8\%} \\
& 0.6   & 52.1 & 9802.5 & 340.5 & 28.8 & 72.2\% \\
& 0.7   & 51.7 & 9719.9 & 339.8 & 28.6 & 71.6\% \\
\bottomrule
\end{tabular}%
}
\caption{Effect of different early stopping thresholds (0.4, 0.5, 0.6, 0.7) on the SFT performance of DeepSeek-R1-Distill-Qwen-7B across mathematical benchmarks.}
\label{tab:threshold}
\end{table}

Observing the distributions in Figure~\ref{fig:earlyratio}, we found that these $t_e$ settings generally cause the model to achieve a confidence score $\hat{c}_i > t_e$ at the position of the first reflection, which aligns with our three guiding principles. Furthermore, the confidence monitored by our designed lightweight detector rarely indicates high confidence before the FAS (Final Answer Serialization) step. This is advantageous as it effectively reduces the computational overhead associated with prematurely judging the correctness of an answer, making it highly compatible with the ConCISE methodology.

Therefore, we further conducted experiments for $t_e \in \{0.4, 0.5, 0.6, 0.7\}$ under the SFT setting of DeepSeek-R1-Distill-Qwen-7B, and Table~\ref{tab:threshold} presents these experimental results. We found that under the settings of $t_e=0.5, 0.6,$ and $0.7$, the model ultimately exhibited similar performance in both compression rate (CR) and accuracy (Acc), demonstrating robustness to the choice of $t_e$. Consequently, we selected $t_e=0.5$ for subsequent extensive experiments. However, when $t_e=0.4$, there was a significant decline in performance. This is because when $t_e$ is set to $0.4$, more data exists at the FAS step, preventing the model from adequately checking its reasoning process; this excessive confidence leads to a performance drop.

\subsection{Reflection step detection}
\label{cha:ref_detect}

The detection of reflection steps, as mentioned throughout this paper, plays a crucial role in both the algorithmic flow and experimental evaluation of ConCISE. Specifically, we employ two distinct methods for identifying these steps, selecting the most suitable one depending on the specific application context. These methods and their corresponding usage scenarios are detailed below.

\subsubsection{Rule-based detection.} 
\label{cha:rule_based_detection}
Reflection steps generated by the model often exhibit distinct linguistic characteristics, frequently containing specific keywords (e.g., \texttt{"wait"}, \texttt{"alternatively"}) that signal potential reasoning shifts or reconsiderations. Consequently, a straightforward and effective strategy for identifying the start of such steps is rule-based detection. This method checks for the presence of predefined "reflection keywords" within the generated text of a given step, $S_i$.

Formally, let $K$ be the predefined set of reflection keywords. A step $s_i$ is identified as the start of a reflection, denoted as $\text{IsReflectionStart}(s_i)$, if its text contains any keyword $k$ from the set $K$:
$$
\label{eq:rule_based_detection} 
\text{IsReflectionStart}(s_i) \equiv \bigvee_{k \in K} (k \subseteq s_i)
$$
where $k \subseteq s_i$ indicates that the keyword $k$ appears as a substring of step $s_i$.The set $K$ of reflection keywords used for rule-based detection consists of the following:
\begin{itemize}
    \item \texttt{"wait"}, \texttt{"alternatively"}, \texttt{"check"}, \texttt{"reconsider"}, \texttt{"reflect"}, \texttt{"rethink"}, \texttt{"reconsidering"}, \texttt{"reviewing"}, \texttt{"reassess"}, \texttt{"pause"}, \texttt{"second thought"}, \texttt{"reevaluate"}, \texttt{"verify"}, \texttt{"think again"}.
\end{itemize}

While effective for identifying the onset of a reflection sequence, this method typically cannot recognize subsequent steps within the same reflection process, as these later reflection steps often lack the initial triggering keywords. Despite this limitation, we utilize this rule-based approach within the \textbf{Confidence Injection}. Since Confidence Injection operates during the model's generation process, identifying and potentially preventing only the first redundant reflection step is often sufficient to avoid generating the entire reflection. This method is particularly advantageous here due to its efficiency (introducing no computational overhead) and ease of implementation.

Similarly, this rule-based detection is used for the Figure~\ref{fig:refprob} analysis to determine if an intervention (like inserting a confidence phrase) triggers the start of a new reflection, as only detecting the initial step with this lightweight method is necessary.

\begin{figure*}[t]
    \centering 

    \begin{subfigure}[b]{0.48\textwidth} 
        \centering
        \includegraphics[width=\linewidth]{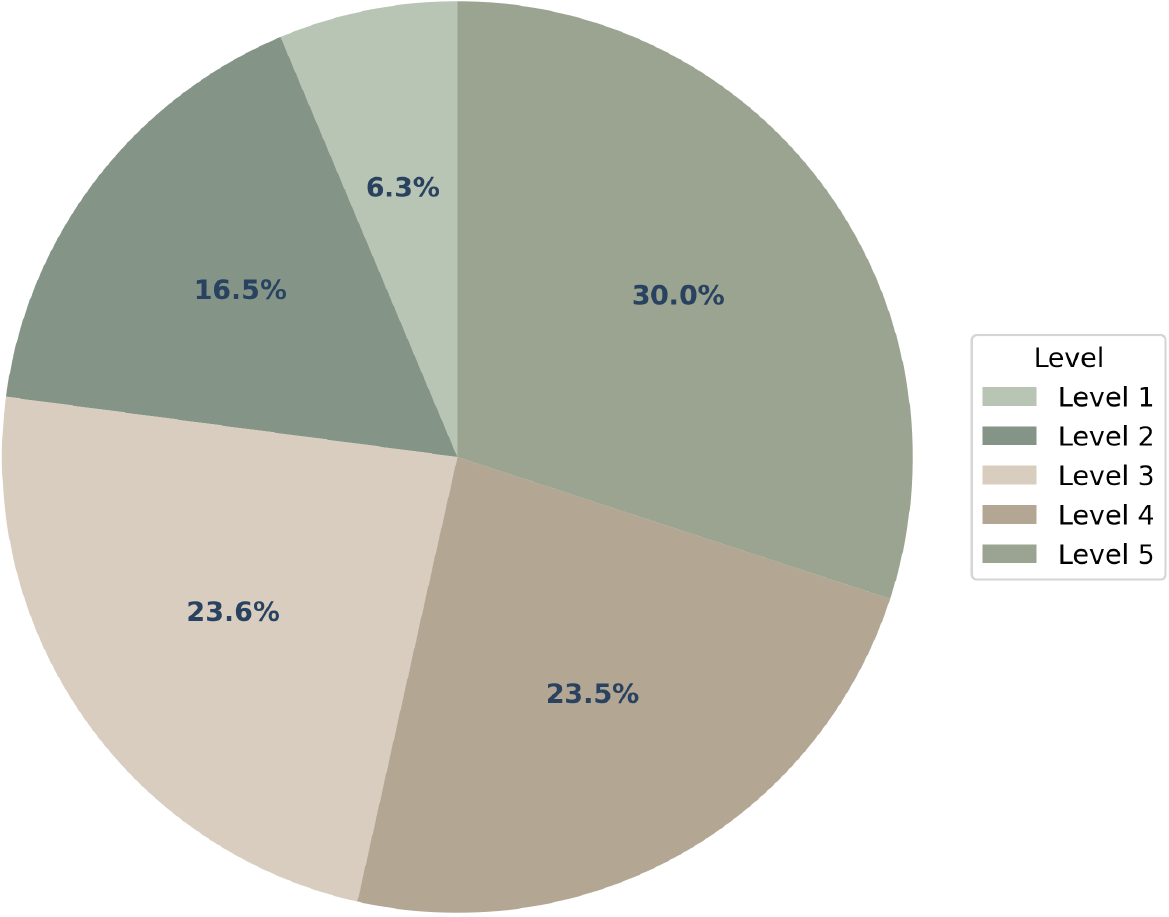}
        \caption{Difficulty diversity of question set.}
        \label{fig:div_dif}
    \end{subfigure}
    \hfill 
    \begin{subfigure}[b]{0.48\textwidth}
        \centering
        \includegraphics[width=\linewidth]{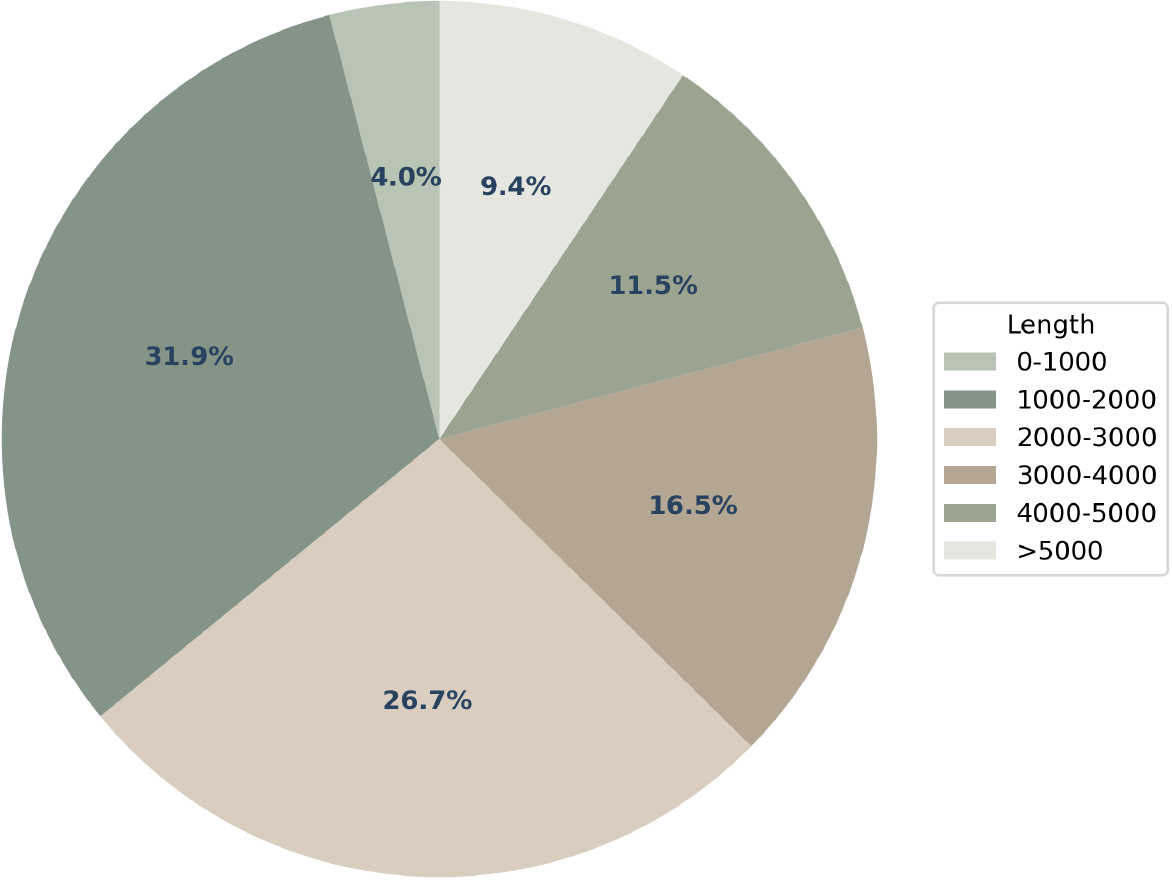}
        \caption{Length diversity of question set.}
        \label{fig:div_len}
    \end{subfigure}

    \caption{Diversity of the question set, showing distributions for difficulty and length.} 
    \label{fig:question_set_diversity} 
\end{figure*}

\subsubsection{LLM-as-a-Judge}
\label{sec:llm_judge} 

When the task requires identifying all reflection steps within an entire reflection behavior—a capability beyond the rule-based method—a more comprehensive approach is needed. For example, implementing the OverThink baseline necessitates isolating and retaining the complete first reflection sequence that occurs after the FAS(First Answer Step). To address this, we employ an LLM-as-a-Judge methodology.

Specifically, we utilize the Qwen-Max model as the judge. We provide it with both the original problem/question and the model's complete reasoning chain as input. The prompt supplied to Qwen-Max includes precise definitions of the FAS and the characteristics defining a 'reflection behavior'. The model is instructed to return a structured output that annotates the input chain, identifying the index of the FAS and providing the indices for all detected reflection steps, grouped according to the reflection behavior they belong to. We then parse this structured output to finalize the detection and grouping of reflection steps. Illustrative examples of the prompt structure, input format, and expected response can be found in Figure~\ref{fig:prompt}, Figure~\ref{fig:input}, and Figure~\ref{fig:output}

Our observations indicate that Qwen-Max performs reliably in identifying FAS and grouping reflection steps when the reasoning chains are of moderate length. However, we noted a decline in performance for chains exceeding 5000 tokens. This degradation is likely attributable to the known limitations of current LLMs in processing very long contexts effectively. Consequently, for the implementation of the OverThink baseline comparison, reasoning chains longer than 3000 tokens were manually annotated to ensure accuracy.

Furthermore, the LLM-as-a-Judge approach is the basis for calculating several reflection-based metrics presented in our results, such as \texttt{RefNum}, \texttt{Non-RefNum}, \texttt{pre-FAS RefNum}, and \texttt{post-FAS RefNum} (shown in Figure~\ref{fig:math_metric} and Figure~\ref{fig:train_metric}). 

\begin{table*}[t]
\centering
\small
\renewcommand{\arraystretch}{1.0}
\resizebox{\linewidth}{!}{%
\begin{tabular}{ll|cc|cc|cc|cc}
\toprule
\multirow{2}{*}{\textbf{Model}} & \multirow{2}{*}{\textbf{Method}} 
& \multicolumn{2}{c|}{\textbf{GSM8K}} 
& \multicolumn{2}{c|}{\textbf{Math-500}} 
& \multicolumn{2}{c|}{\textbf{AIME24}} 
& \multicolumn{2}{c}{\textbf{Average}} \\
\cmidrule{3-10}
& & Acc. & CR & Acc. & CR & Acc. & CR & Acc. & CR \\
\midrule
\multirow{5}{*}{DeepSeek-7B} 
& Origin & 93.1 & 100\% & 90.8 & 100\% & 54.2 & 100\% & 79.4 & 100\% \\
& DEER(0.98) & 90.3 & 59\% & 87.0 & 51\% & 47.1 & 69\% & 74.8 & 60\% \\
& DEER(0.95) & 87.9 & 56\% & 85.2 & 49\% & 43.4 & 64\% & 72.2 & 56\% \\
& TALE & 89.0 & 47\% & 90.0 & 95\% & 52.9 & 98\% & 77.3 & 80\% \\
& ConCISE-Decoding & 92.7 & 62\% & 89.2 & 55\% & 49.6 & 67\% & 77.2 & 61\% \\
& \textbf{ConCISE-SimPO} & 92.1 & \textbf{50\%} & 91.0 & \textbf{51\%} & \textbf{48.3} & \textbf{57\%} & 77.1 & \textbf{53\%} \\
\midrule
\multirow{5}{*}{DeepSeek-1.5B} 
& Origin & 85.4 & 100\% & 82.2 & 100\% & 29.2 & 100\% & 65.6 & 100\% \\
& DEER(0.98) & 79.7 & 41\% & 77.6 & 54\% & 24.6 & 70\% & 60.6 & 55\% \\
& DEER(0.95) & 74.7 & 43\% & 76.4 & 49\% & 21.3 & 64\% & 57.5 & 52\% \\
& TALE & 84.9 & 90\% & 81.6 & 98\% & 29.6 & 98\% & 65.4 & 95\% \\
& ConCISE-Decoding & 83.9 & 53\% & 79.0 & 52\% & 26.3 & 52\% & 63.1 & 52\% \\
& \textbf{ConCISE-SimPO} & 84.3 & \textbf{36\%} & 83.6 & \textbf{51\%} & \textbf{30.4} & \textbf{50\%} & 66.1 & \textbf{46\%} \\
\bottomrule
\end{tabular}%
}
\caption{Detailed results of training-free baselines and \textsc{ConCISE} variants on DeepSeek-R1-Distill-Qwen-7B and 1.5B across GSM8K, Math-500, and AIME24. 
Acc: accuracy (\%), CR: compression ratio relative to original (lower CR indicates shorter outputs).}
\label{tab:trainingfree}
\end{table*}

\begin{figure*}[htbp]
\centering
\begin{subfigure}[htbp]{0.32\linewidth}
    \centering
    \includegraphics[width=\linewidth]{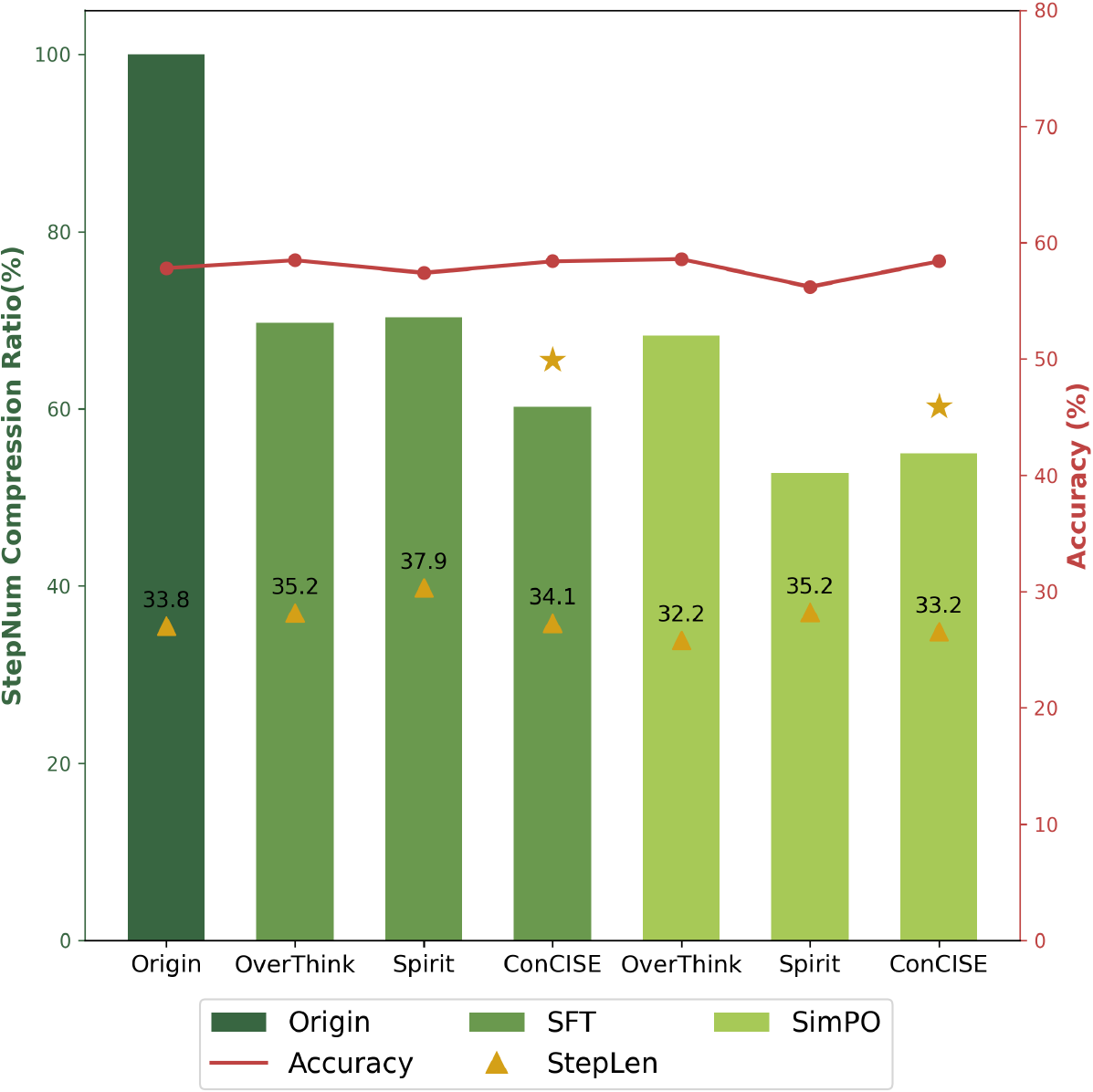}
    \caption{Average \texttt{Acc}, \texttt{StepNum}, and \texttt{StepLen}.}
    \label{fig:steplen_1.5B}
\end{subfigure}
\hfill
\begin{subfigure}[htbp]{0.3\linewidth}
    \centering
    \includegraphics[width=\linewidth]{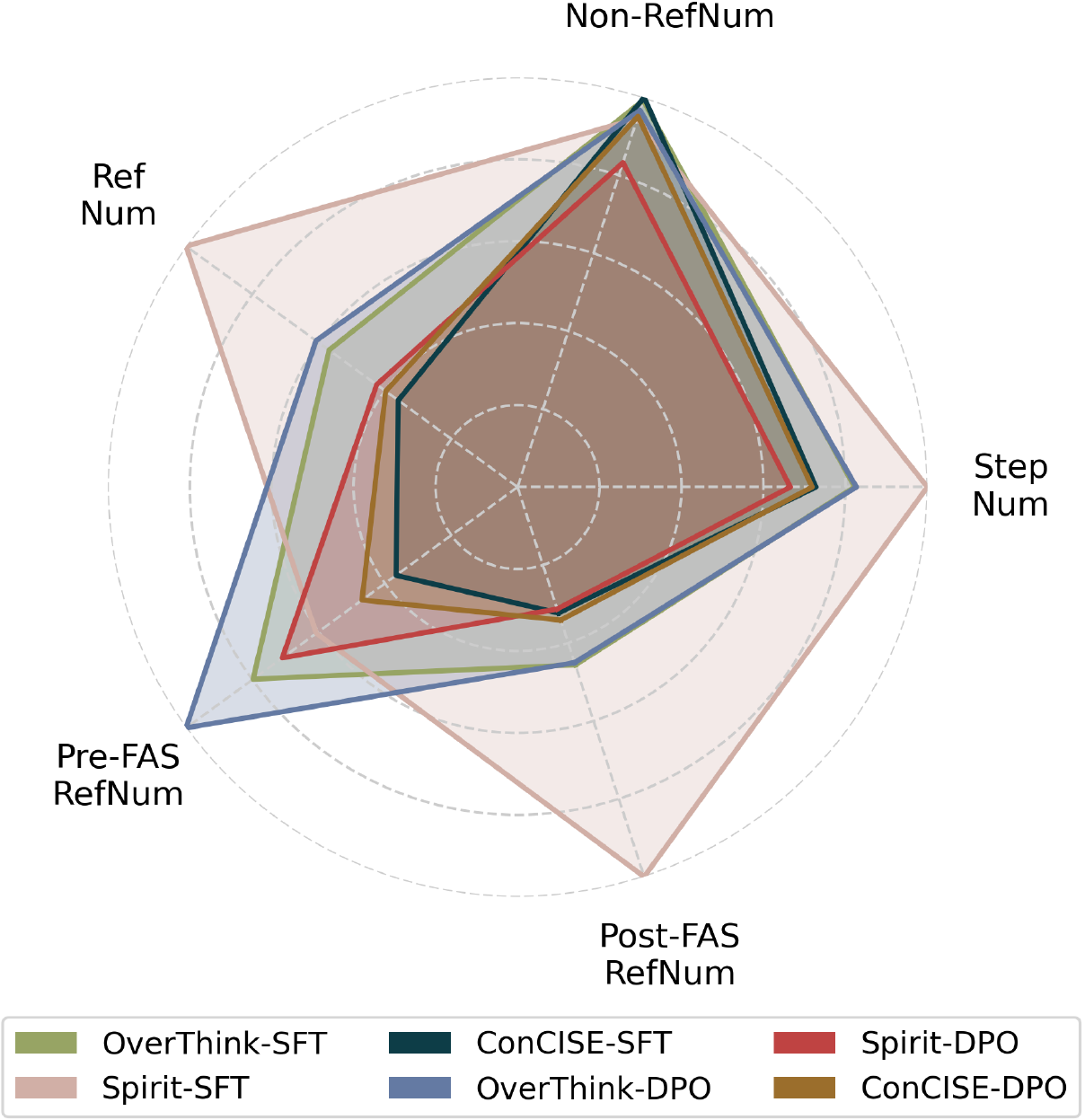}
    \caption{Analysis of Reflection steps.}
    \label{fig:math_metric_1.5B}
\end{subfigure}
\hfill
\begin{subfigure}[htbp]{0.3\linewidth}
    \centering
    \includegraphics[width=\linewidth]{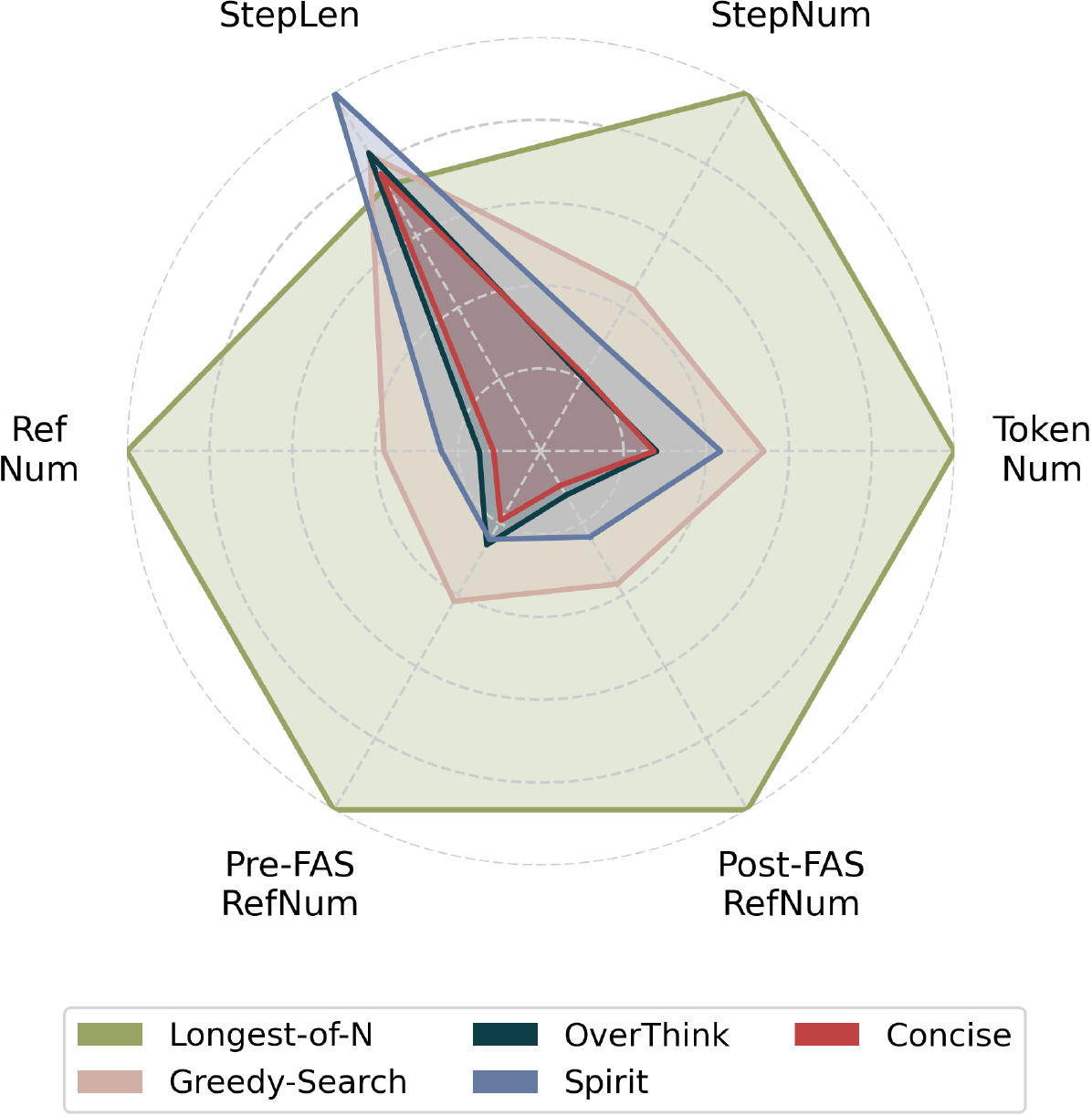}
    \caption{Metrics of training datasets.}
    \label{fig:train_metric_1.5B}
\end{subfigure}
\caption{Further analysis of reasoning chain and training dataset characteristics on DeepSeek-R1-Distill-Qwen-1.5B.}
\label{fig:analysis_1.5B}
\end{figure*}

\section{Details about Experiments}
\label{cha:experiments_datails}
\subsection{Datasets}
\label{cha:datasets_details}

\paragraph{Question Set}
While the construction methodology for the question set is detailed in Section~\ref{sec:exp}, here we focus on illustrating its diversity in terms of difficulty and length.Figures~\ref{fig:div_dif} and \ref{fig:div_len} demonstrate the diversity of the Question set. Difficulty distribution is derived from metric \texttt{Level} in the MATH dataset, while length distribution reflects reasoning chain lengths from DeepSeek-R1-Distill-Qwen-7B under greedy decoding. Furthermore, it is important to note a subsequent refinement process for these selected questions when constructing the final training set. Although the initial 2000 data entries were chosen based on the premise that the model could correctly answer them using greedy search, the generation processes for our \textsc{ConCISE} and OverThink methods do not strictly adhere to greedy decoding, and \textsc{ConCISE} will discard wrong reasoning chains. Therefore, for training purposes, we ultimately utilize the common subset of correctly reasoned chains derived from the outputs of three methods: \textsc{ConCISE}, OverThink, and Spirit. This intersection results in a refined training dataset of approximately 1900 samples.

\paragraph{Validation Dataset}
For the experiments presented in Figure~\ref{fig:verify}, we curated the Verification Dataset. This was necessary because the required analyses (including First Answer Step (FAS) and reflection detection, as shown in Figure~\ref{fig:confidence}) rely on annotations from the LLM-as-a-Judge method (\S\ref{sec:llm_judge}), whose reliability decreases on long reasoning chains. To ensure dependable results, the Verification Dataset comprises another 1000 reasoning chains selected from the MATH training set, filtered based on two criteria: yielding a correct answer under greedy decoding and having a reasoning chain length under 5000. This curated set facilitates reliable LLM-based annotation for the validation experiments.

\paragraph{Evaluation Dataset}
The analysis presented in Figure~\ref{fig:math_metric} and Figure~\ref{fig:math_metric_1.5B} required LLM-as-a-Judge annotations. To ensure reliable results despite this method's limitations with long contexts, we filtered the MATH500 dataset. We selected only those problems where reasoning chains generated by all twelve specified finetuning configurations (combinations of ConCISE/OverThink/Spirit methods, 1.5B/7B sizes, and SFT/SimPO techniques) were shorter than 5000 tokens. This yielded a common subset of 340 problems, on which our subsequent comparative analysis of the different methods was exclusively performed.

\subsection{Training Details}
\label{cha:training_details}

\paragraph{Models.} DeepSeek-R1-Distill-Qwen-7B and DeepSeek-R1-Distill-Qwen-1.5B are reasoning models derived by DeepSeek from the Qwen2.5 series models  \cite{qwen2.5}through distillation training, using training and generation data from the DeepSeek-R1 model. Skywork-OR1-7B-Preview was further developed via reinforcement learning, exhibiting strong capabilities on test sets such as mathematics. Meanwhile, Qwen3-8B is a recently released hybrid reasoning model noted for its powerful reasoning abilities. All four of these are open-source models. While they possess strong Chain-of-Thought (CoT) reasoning capabilities, they also present clear issues with reasoning redundancy. Consequently, we employ these four models for subsequent training and evaluation. Furthermore, we ensure that these models, along with all datasets involved in the training and evaluation processes, are utilized following their original licenses and intended purposes.

\paragraph{Training.} We construct training datasets based on the question set using \textsc{ConCISE}, OverThink, and Spirit. For the generation process of ConCISE, we set Temperature=0.6, Top\_p=0.95. For OverThink, we sample 8 reasoning chains under Tempature=1.0(which is relatively high due to the recommended temperature for solving math problems of the LRMs is usually 0.6-0.8), Top\_p=0.95. For Spirit, compression is applied to reasoning chains generated via greedy search. During training, we adopt two strategies: SFT and SimPO. SFT directly uses the question and its corresponding compressed reasoning trace for supervised fine-tuning. For SimPO, reject samples are also required \cite{rafailov2023direct}. Following the original OverThink setup, we use the longest correct reasoning chain among the eight samples as the reject samples. To ensure a fair comparison, \textsc{ConCISE} adopts the same strategy. For Spirit, the full reasoning chain before compression serves as reject samples, because this is precisely the comparative effect Spirit aims to demonstrate: important steps are retained while preference relationships are learned only for non-important steps. We use LlamaFactory for both SFT and SimPO fine-tuning \cite{zheng2024llamafactory}. All models are trained with $lr=1e-6$; SFT runs for 2 epochs and SimPO for 1 epoch. The effective batch size for all training is 32 (using \texttt{per\_device\_train\_batch\_size=1} and \texttt{gradient\_accumulation\_steps=8} on 4 A800 GPUs), and \texttt{max\_length} is set to 10240.

\subsection{Analysis on DeepSeek-1.5B}
\label{cha:analysis_1.5B}

Regarding the analysis conducted on DeepSeek-R1-Distill-Qwen-7B in the Experimental Section~\ref{sec:analysis}, we also performed a similar analysis for DeepSeek-R1-Distill-Qwen-1.5B. Figure~\ref{fig:analysis_1.5B} shows the specific results. The trends exhibited by the data, as well as the relationships among the three methods, are largely consistent with those presented for DeepSeek-R1-Distill-Qwen-7B. Consequently, the final conclusions are also consistent.

\clearpage
\begin{figure*}[htbp]
\centering
\begin{tcolorbox}[title=Prompt, width=\textwidth]
You are an AI assistant trained to analyze reasoning steps in a response. Your task has two parts:\\
1. Examine each reasoning step to determine if it's part of a reflection process.\\
2. Identify the earliest step where the final answer (as later shown in boxed\{\}) is first derived, regardless of whether it is formally written or boxed at that moment.

\vspace{1em}
\textbf{\#\#\# [Definition of Reflection]}:\\
1. A reflection process is a sequence of one or more reasoning steps that recheck or doubt a previously made conclusion, such as double-checking calculations, using alternative methods.\\
2. Typical signals include (but are not limited to): ‘Wait’, ‘Alternatively’, ‘Just to double check’, 'But hold on', etc. These signals usually mean the start of a new reflection process.\\
3. However, even without such phrases, if the content of a step reflects a verification or reevaluation, it should be marked as a part of a reflection process.

\vspace{1em}

\textbf{\#\#\# [Output Format]}:\\
1. \textbf{Reflection Step}: List all reflection processes as groups of steps.\\
    \hspace*{1.5em} - If Step3 and Step4 form a reflection, write as (Step3, Step4)\\
    \hspace*{1.5em} - If Step5, Step6, Step7 form a new reflection process together, list as a separate group: [(Step3, Step4), (Step5, Step6, Step7)]\\
    \hspace*{1.5em} - Avoid putting a lot of steps into one single reflection process.\\
2. \textbf{First Answer Step}: Write the earliest step where the final answer is first derived(e.g., Step2).
            
\vspace{1em}
\textbf{\#\#\# [Example]}:\\
        \textbf{Question}: 2 + 3 = ? \\
        \textbf{Response}:
            \textit{Step1}: I start with 2 + 3.\;
            \textit{Step2}: That gives me 5.\;
            \textit{Step3}: Wait, is that right? Let me make sure...\;
            \textit{Step4}: But hold on, let me double-check. Maybe I should...\;
            \textit{Step5}: Wait, no, I think...\; 
            \textit{Step6}: Therefore, the result is 5\;
            \textit{Step7}: Alternatively, if I use...\; 
            \textit{Step8}: So, the answer is 5\;
            \textit{Step9}: **Final Answer**: the answer is boxed{5}.\\
        \textbf{Output}:\\
            Reflection Step: [(Step3), (Step4), (Step5, Step6), (Step7, Step8)]\\
            First Correct Step: Step2\\      
        \textbf{Explanations for Reflection Step}:\\
            \hspace*{1.5em}- There is 'wait', a reflection keyword, in Step3, so Step3 is the start of a reflection process. \\
            \hspace*{1.5em}- There is 'But hold on' in step4, so step4 means the start of a new reflection process. \\
            \hspace*{1.5em}-  Then, there is 'wait' in step5, so step5 means the start of a new reflection process.\\ 
            \hspace*{1.5em}- There is 'therefore' in step6, so (step5,step6) is in the same refleciton process.\\
            \hspace*{1.5em}- There is 'alternatively' in step7, so step7 is the start of a new reflection process.\\
            \hspace*{1.5em}- There is 'so' in step8, so (step7,step8) is in the same reflection process.
            
\vspace{1em}
\textbf{\#\#\# [Task]}:\\
Now analyze the following question and response:\\
f"Question: \textcolor[rgb]{0.5,0.0,0.0}{\{question\}}"\\
f"Response: \textcolor[rgb]{0.5,0.0,0.0}{\{response\}}"\\
Please output the reflection steps and the first correct step in the format without extra explanation.

\vspace{1em}
\end{tcolorbox}
\caption{Prompt of using the LLM-as-a-Judge method for reflection step detection.}
\label{fig:prompt}
\end{figure*}

\begin{figure*}[htbp]
\centering
\begin{tcolorbox}[title=Input, width=\textwidth]
\textbf{Question}:
\begin{itemize}
    \item What day of the week will it be 40 days from Tuesday? 
\end{itemize}

\textbf{Response}:
\begin{itemize}
    \item Step1: Alright, so I need to figure out what day of the week it will be 40 days from Tuesday. Hmm, okay. Let me think about how to approach this.
    \item Step2: First, I remember that each week has 7 days. So, if something happens every 7 days, it's on the same day of the week next time. That might be useful here. Maybe I can figure out how many weeks are in 40 days and then see how many extra days that is.
    \item Step3: Let me do the division: 40 divided by 7. I think that's 5 weeks and 5 days.
    \item Step4: \textcolor[rgb]{0.0,0.2,0.6}{\textbf{Wait, let me check.}} 7 times 5 is 35, right? So, 40 minus 35 is 5. So, 40 days is equal to 5 weeks and 5 days. That makes sense.
    \item Step5: So, if today is a Tuesday, then 35 days later is also a Tuesday. Because 35 is exactly 5 weeks, which is 5 times 7. So, after 35 days, it's still Tuesday. Now, I need to find out what day it is 5 days after Tuesday.
    \item Step6: Let me count: Tuesday plus one day is Wednesday, plus two days is Thursday, plus three days is Friday, plus four days is Saturday, and plus five days is Sunday. So, 5 days after Tuesday is  \textcolor[rgb]{0.5,0.0,0.0}{\boxed{Sunday}}.
    \item Step7: Therefore, 40 days from Tuesday is Sunday. Let me \textcolor[rgb]{0.0,0.2,0.6}{\textbf{double-check}} that to make sure I didn't make a mistake. If today is Tuesday and 7 days later is also Tuesday, then adding another 7 days each time cycles through the week. So, 35 days later, as we calculated, it is Tuesday again. Then, 5 days beyond that would be Sunday. Yeah, that seems right.
    \item Step8: \textbf{Final Answer}: The day of the week 40 days from Tuesday is \textcolor[rgb]{0.5,0.0,0.0}{\boxed{Sunday}}.
\end{itemize}

\vspace{1em}
\end{tcolorbox}
\caption{An example input of using the LLM-as-a-Judge method for reflection step detection.}
\label{fig:input}
\end{figure*}

\begin{figure*}[htbp]
\centering
\begin{tcolorbox}[title=Output, width=\textwidth]
\textbf{Output}:
\begin{itemize}
    \item Reflection Step: [(Step4), (Step7)]
    \item First Answer Step: Step6
\end{itemize}
\end{tcolorbox}
\caption{An example output of using the LLM-as-a-Judge method for reflection step detection.}
\label{fig:output}
\end{figure*}

\end{document}